\documentclass{article}

   \usepackage[preprint]{neurips_2024}

\usepackage[utf8]{inputenc} % allow utf-8 input
\usepackage[T1]{fontenc}    % use 8-bit T1 fonts
\usepackage{hyperref}       % hyperlinks
\usepackage{url}            % simple URL typesetting
\usepackage{booktabs}       % professional-quality tables
\usepackage{amsfonts}       % blackboard math symbols
\usepackage{nicefrac}       % compact symbols for 1/2, etc.
\usepackage{microtype}      % microtypography
\usepackage{xcolor}         % colors
\usepackage{amsmath}
\usepackage{amssymb}
\usepackage{mathtools}
\usepackage{amsthm}
\usepackage{graphicx}
\usepackage{algorithm}
\usepackage{algpseudocode}
\usepackage{multirow}
\usepackage{adjustbox}

\usepackage{xparse}

\title{Curvature Clues: Decoding Deep Learning Privacy with Input Loss Curvature}

\DeclareMathOperator{\Ex}{\mathop{\mathbb{E}}}
\DeclareMathOperator{\Curv}{\mathrm{Curv}}
\DeclareMathOperator{\KL}{\mathrm{D_{KL}}}
\DeclareMathOperator{\tr}{\mathrm{tr}}
\DeclarePairedDelimiterX{\set}[1]{\{}{\}}{\setargs{#1}}\NewDocumentCommand{\setargs}{>{\SplitArgument{1}{;}}m}{\setargsaux#1}
\NewDocumentCommand{\setargsaux}{mm}
{\IfNoValueTF{#2}{#1} {#1\,\delimsize|\,\mathopen{}#2}}

\theoremstyle{plain}
\newtheorem{theorem}{Theorem}[section]

\newtheorem{lemma}[theorem]{Lemma}

\theoremstyle{definition}

\theoremstyle{remark}

\author{%
  Deepak Ravikumar\quad Efstathia Soufleri\quad  Kaushik Roy\\
  Department of Electrical and Computer Engineering\\
  Purdue University\\
  West Lafayette, IN 47906 \\
  \texttt{\{dravikum, esoufler, kaushik\}@purdue.edu} \\
}

\begin{document}

\maketitle

\begin{abstract}
    In this paper, we explore the properties of loss curvature with respect to input data in deep neural networks. Curvature of loss with respect to input (termed input loss curvature) is the trace of the Hessian of the loss with respect to the input. We investigate how input loss curvature varies between train and test sets, and its implications for train-test distinguishability. We develop a theoretical framework that derives an upper bound on the train-test distinguishability based on privacy and the size of the training set. This novel insight fuels the development of a new black box membership inference attack utilizing input loss curvature. We validate our theoretical findings through experiments in computer vision classification tasks, demonstrating that input loss curvature surpasses existing methods in membership inference effectiveness. 
    % Moreover, our analysis also sheds light on the potential of using subsets of training data as a defense mechanism against shadow model based membership inference attacks, revealing a previously unknown limitation of shadow model based methods. 
    Our analysis highlights how the performance of membership inference attack (MIA) methods varies with the size of the training set, showing that curvature-based MIA outperforms other methods on sufficiently large datasets. This condition is often met by real datasets, as demonstrated by our results on CIFAR10, CIFAR100, and ImageNet.
    These findings not only advance our understanding of deep neural network behavior but also improve the ability to test privacy-preserving techniques in machine learning.
\end{abstract}

% \begin{abstract}
%     In this paper we study the properties of loss curvature with respect to the input. Specifically, we study loss curvature with respect to the input (a.k.a input loss curvature) and how it relates to train and test sets in deep neural nets. We build a theoretical framework to study the distribution of input loss curvature scores for samples in the train set versus samples in the test set. Our theoretical analysis provides an upper bound on the train-test distinguishability.  The results of which suggest that the train-test distinguishability when using input loss curvature is upper bound by the number of samples in the training set. Following this new insight leads us to develop a new black box membership inference attack and defense methodology using input loss curvature. We perform experiments targeting computer vision classification tasks to validate the theoretical results and establish the superiority of using input loss  curvature for membership inference by comparing against state-of-the-art methods. Further, the theoretical results also provide new insights into the use of input loss curvature based coresets for privacy utility trade-off. 
% \end{abstract}

\section{Introduction}

% # don;t start with privacy start with curvature how it related to genralization, # weights but not really input, only been studeid for adversari robsutness. 
Deep neural networks are being increasingly trained on sensitive datasets; thus ensuring the privacy of these models is paramount. Membership inference attacks (MIA) have become the standard approach to test a model's privacy \citep{murakonda2020ml}. These attacks take a trained model and aim to identify if a given example was used in its training. Recent work has linked curvature of loss with respect to input with memorization \citep{garg2023memorization} and differential privacy \citep{dwork2006calibrating, ravikumar2024unveiling}. Inspired by this line of research, we investigate the properties of input loss curvature and leverage our insights to develop a new membership inference attack.
% and defense technique.

% study one of the fundamental properties of deep neural nets, specifically the curvature of the loss landscape with respect to the input data. 

Curvature of loss with respect to input (termed input loss curvature) is defined as the trace of the Hessian of loss with respect to the input \citep{moosavi2019robustness, garg2023samples}.  % We investigate it for its variation between train and test sets and its implications for train-test distinguishability. 
Prior works that study loss curvature have focused on two lines of research. The first line of research has focused on studying the loss curvature with respect to the weights of the deep neural net \citep{keskar2017on, wu2020adversarial, jiang2020fantastic, foret2021sharpnessaware, kwon2021asam, andriushchenko2022towards} to better understand generalization. The second line of research studied the loss curvature with respect to the input (i.e. data) for gaining insight into adversarial robustness \citep{moosavi2019robustness, fawzi2018empirical}, coresets \citep{garg2023samples} and memorization \citep{garg2023memorization, ravikumar2024unveiling}, mainly focusing on its properties on the train set.  Thus, there is a gap in our understanding of input loss curvature on unseen test examples. 

\begin{figure}
    \centering
    \includegraphics[width=0.9\linewidth]{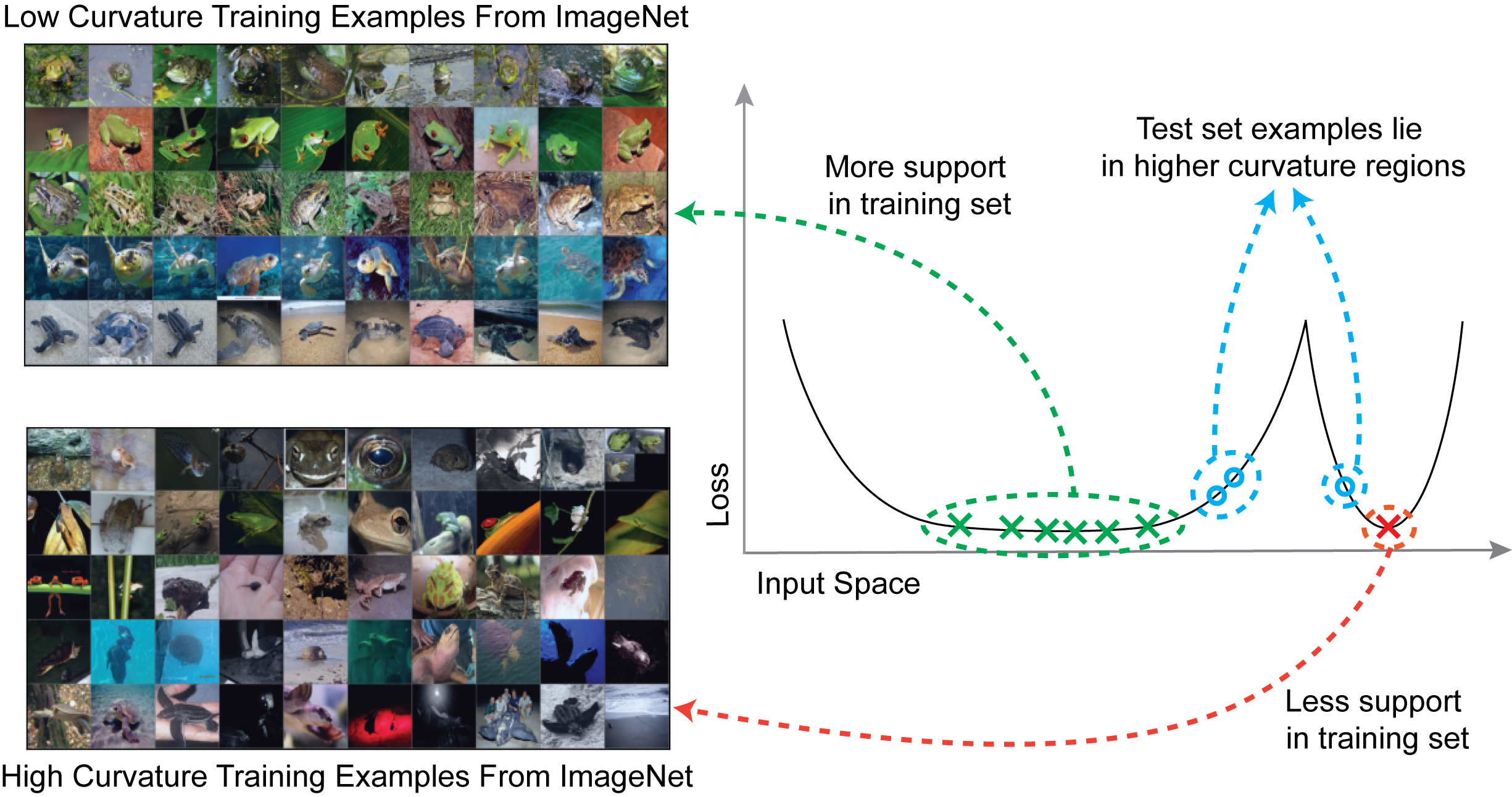}
    \caption{Visualizing low and high input curvature samples from a ResNet50 trained on ImageNet. Low input curvature training set images are prototypical and have lots of support in the trainset, while high input curvature train set examples have less support and are atypical. Test set examples lie around the training set images in higher curvature regions.}
    \label{fig:intro}
    % \vspace{-4mm}
\end{figure}

% In this paper we study the properties of input loss curvature obtained on the test set. 
Input loss curvature on the train set captures the proto-typicality of an image. This is visualized in Figure \ref{fig:intro} (left) which shows high curvature examples from ImagetNet \citep{ILSVRC15} train set. Figure \ref{fig:intro} (right) builds intuition as to why this is the case -- a sample with lots of support in the dataset is more likely to lie in low curvature regions while atypical/non-prototypical examples have less support in the training set and thus lie in higher curvature regions \citep{garg2023memorization, ravikumar2024unveiling}. With this established,  we make a novel observation that on average the input loss curvature scores for test set examples are higher than train set examples. This is because test samples were not optimized for, hence they lie slightly off the flat minima, in regions of higher input curvature (also visualized in Figure \ref{fig:intro}). 
% This distinguishability between train-test examples motivates us to build a theoretical framework to study test set input loss curvature. 
We leverage this insight and develop a theoretical framework that focuses on the distinguishability of train-test input loss curvature scores. Studying this distinguishability leads us to understand performance bounds of membership inference attacks \citep{shokri2017membership, sablayrolles2019white, song2021systematic, carlini2022membership}.

Our theoretical analysis obtains an upper bound on the KL divergence between train-test input loss curvature scores. This provides a ceiling for membership inference attack performance. The analysis also reveals that the upper bound when using input loss curvature scores is dependent on the number of training set examples and the differential privacy parameter $\epsilon$. This insight helps us understand the conditions under which input loss curvature can be more effective at detecting train vs. test examples. Conditions such as the number of training examples and the privacy guarantees of the model.

% This insight leads us to gain insights on the conditions (i.e. when is the number of train examples large enough? How does privacy affect performance?) under which input loss curvature can be be more sensitive at detecting train vs. test examples.  

To test our theoretical results 
% and investigate if curvature scores perform better at larger training set sizes, 
we perform black-box (setting where adversaries have access only to model outputs) membership inference attack (MIA) using input loss curvature scores.
However, input loss curvature score calculation needs access to model parameters (which are not available in a black box setting). To resolve this issue we propose using a zero-order input loss curvature estimation. Zero-order estimation can calculate input loss curvature without needing access to model parameters. 
The results of our input loss curvature-based black-box membership inference attack demonstrate that real datasets have a sufficient number of training examples for curvature scores to achieve superior performance compared to existing state-of-the-art membership inference techniques.
% The result of input loss curvature based black-box membership inference attack shows that real datasets have sufficient number of training set examples for curvature scores to yield superior performance compared to existing state-of-the-art membership inference techniques. 
% Further, the dependence of train-test distinguishability on the number of samples leads us to study the performance of membership inference attacks on the size of the training set.
Our results show that curvature based MIA outperforms prior state-of-the-art techniques on 10\% subset of CIFAR100 dataset (about 5000 sample training set). 
% Results show that training on smaller subsets of the dataset may provide defense against shadow model based MIA, thus revealing a previously unknown limitation of shadow model based MIA.
% Results show that curvature-based MIA outperforms other methods on sufficiently large datasets. This condition is often met by real datasets, as demonstrated by our results on CIFAR-10, CIFAR-100, and ImageNet.

In summary, our contributions are as follows:
\begin{itemize}
\item Theoretical Foundation: We provide theoretical analysis to understand the train-test distinguishability with input loss curvature scores, demonstrating that input loss curvature is more sensitive and hence better at detecting train vs. test set examples than current state-of-the-art membership inference attacks.
\item Adapting Theory To Practice: We propose using zero order input loss curvature estimation to enable black box membership inference attack using input loss curvature scores.
\item Better Attack:  We conduct experiments to validate our theoretical results. Specifically, we show that input loss curvature enables more effective black box membership inference attacks when compared to existing state-of-the-art techniques.
% \item  Improved Privacy Testing:  Guided by the theoretical insights, we study how the number of training examples affects membership inference attack performance. We show that training on subsets of the dataset may provide defense against shadow model based MIA, revealing a previously unknown limitation of shadow model based methods.
% Thus, our recommendation for ML practitioners testing privacy is that, they should evaluate privacy using multiple MIA methods and include methods without shadow models for a comprehensive test.
\end{itemize}
\vspace{-2mm}
\section{Related Work}

% \textbf{Memorization} has garnered increasing research effort with several recent works aiming to add to the understanding of memorization and its implications \cite{zhang2017understanding, arpit2017closer, carlini2019distribution, feldman2019high, feldman2019does, feldman2020neural, maini2022characterizing, garg2023memorization, lukasik2023larger}. The motivation for studying memorization stems from a variety of goals ranging from deriving insights into generalization \cite{zhang2017understanding, toneva2018an, brown2021memorization, zhang2021understanding}, identifying mislabeled examples \cite{pleiss2020identifying, maini2022characterizing}, and identifying challenging or rare sub-populations \cite{carlini2019distribution}, to understanding privacy \cite{feldman2019does} and robustness risks from memorization \cite{shokri2017membership, carlini2022membership}. While several metrics have been proposed to study memorization \cite{carlini2019distribution, jiang2020characterizing}, the stability-based memorization score proposed by \citet{feldman2019does} provides a strong theoretical framework to understand memorization along with strong empirical evidence \cite{feldman2020neural}. However, since the score proposed by \citet{feldman2019does} is computationally expensive,   \citet{garg2023memorization} proposed using input loss curvature as a more compute-efficient proxy. In this paper, we develop the theoretical framework to understand the links between input loss curvature, memorization, and differential privacy.

\textbf{Membership Inference Attacks} are used as a tool to test privacy \citep{murakonda2020ml}. These attacks aim to identify if a particular data point was included in a model's training dataset \citep{shokri2017membership}. 
% A successful attack can reveal sensitive information about individuals or organizations.  
Existing techniques often leverage model outputs like loss values \citep{shokri2017membership, yeom2018privacy, sablayrolles2019white}, confidence scores \citep{carlini2022membership} or modified entropy \citep{song2021systematic}. 
 \cite{shokri2017membership} proposed the used of shadow models which are auxiliary models trained on subsets of the target model's data to aid in inference. Several modifications to this approach have been proposed. One important addition is the focus on example hardness, where the authors \citet{sablayrolles2019white} proposed an attack that scaled the loss using per example hardness threshold which is estimated by training shadow models. \citet{watson2021importance} proposed a similar approach but in the offline case where they calibrated the example hardness using the average loss of shadow models not trained on the target example. \citet{ye2022enhanced} models various attacks into four categories and performs differential analysis to explain the gaps between them. Both \citet{ye2022enhanced} and \citet{long2020pragmatic} consider the entire loss distribution of samples that are not in the training set. However, they face challenges in extrapolating to low false positive rates (FPRs). To address this issue \citet{carlini2022membership} propose using a parametric model along with shadow models to improve performance. 
Orthogonal to these approaches \citet{choquette2021label} suggests the use of input augmentations during evaluation to improve the performance of the attacks. Similarly \citet{jayaraman2021revisiting} propose the MERLIN attack, which queries the target model 
multiple times on a sample perturbed with Gaussian noise.

\textbf{Input Loss Curvature} is defined as the trace of the Hessian of loss with respect to the input \citep{moosavi2019robustness, garg2023samples}. The aim is to measure the sensitivity of the deep neural network to a specific input. In general, loss curvature with respect to the parameters of deep neural nets has received lots of attention \citep{keskar2017on, wu2020adversarial, jiang2020fantastic, foret2021sharpnessaware, kwon2021asam, andriushchenko2022towards}, specifically due to its role in characterizing the sharpness of the learning objective which is closely connected to generalization. 
However, loss curvature with respect to the input data has received much less attention. It has been studied in the context of adversarial robustness \citep{fawzi2018empirical, moosavi2019robustness}, coresets \citep{garg2023samples}. It has recently been linked with memorization \citep{garg2023memorization, ravikumar2024unveiling} and privacy \citep{ravikumar2024unveiling}.
The authors in \citet{moosavi2019robustness} showed that adversarial training decreases the input loss curvature and provided a theoretical link between robustness and curvature. In an orthogonal direction, \citet{garg2023samples} proposed the use of low input loss curvature examples as training dataset sets called coresets, which they showed to be more data-efficient. However, all of these works have focused on input loss curvature on the trainset. In this paper we focus on input loss curvature and its behavior on test or unseen examples to understand and improve the performance of membership inference attacks.
% In this paper we explore a input loss curvature a metric other than confidence scores, loss values or entropy
Before we discuss our contributions, we present a few preliminaries, notation, and background needed for this paper
\vspace{-2mm}
\section{Notation and Background}
\label{Preliminaries and Notation}

\textbf{Notation.} Let us consider a supervised learning problem, where the goal is to learn a mapping from an input space $\mathcal{X} \subset \mathbb{R}^d$ to an output space $\mathcal{Y}\subset \mathbb{R}$. The learning is performed using a randomized algorithm $\mathcal{A}$ on a training set $S$. Note that a randomized algorithm employs a degree of randomness as a part of its logic. 
The training set $S$ contains $m$ elements denoted as $z_1, \cdots, z_m$.
% drawn from an unknown distribution $\mathcal{D}$. 
Each element $z_i = (x_i, y_i)$ is drawn from an unknown distribution $\mathcal{D}$, where $z_i \in \mathcal{Z}, x_i \in \mathcal{X}$, $y_i \in \mathcal{Y}$ and $\mathcal{Z} = \mathcal{X} \times \mathcal{Y}$. Thus, we define the training set $S \in \mathcal{Z}^m$ as $S = \{z_1, \cdots, z_m\}$.
% \begin{align*}
%    S = \{z_1, \cdots, z_m\}.
% \end{align*}
% We have the assumption that $m \geq 2$. 
Another related concept is that of adjacent datasets, which are obtained when the set's $i^{th}$ element is removed and defined as
\begin{align*}
    S^{\setminus i} = \{z_1, \cdots, z_{i-1}, z_{i+1}, \cdots, z_m\}.
\end{align*}

Additionally, the concept of adjacent (or neighboring) datasets is linked to the distance between datasets. The distance between two datasets $S$ and $S'$, denoted by $\lVert S - S' \rVert_1$, measures the number of samples that differ between them. The notation $\lVert S \rVert_1$ represents the size of a dataset $S$. When a randomized learning algorithm $\mathcal{A}$ is applied to a dataset $S$, it produces a hypothesis denoted by $h_{S}^\phi = \mathcal{A}(\phi, S)$,
where $\phi \sim \Phi$ is the random variable associated with the algorithm's randomness. A cost function $c:\mathcal{Y} \times \mathcal{Y} \rightarrow \mathbb{R}^+$ is used to measure the hypothesis's performance. The cost of the hypothesis $h$ at a sample $z_i$ is also referred to as the loss $\ell$ at $z_i$, defined as $\ell(h, z_i) = c(h(x_i), y_i)$.

\textbf{Differential Privacy.} Introduced by \citet{dwork2006calibrating}, differential privacy is defined as follows: A randomized algorithm $\mathcal{A}$ with domain $\mathcal{Z}^m$ is considered $\epsilon$-differentially private if for any subset $\mathcal{R} \subset \mathrm{Range} (\mathcal{A})$ and for any datasets $S, S' \in \mathcal{Z}^m$ differing in at most one element (i.e. $||S-S'||_1 \leq 1$):
\begin{align}
\Pr_\phi[h^\phi_S \in \mathcal{R}] & \leq e^{\epsilon} \Pr_\phi[h^\phi_{S'} \in \mathcal{R}],
\label{eq:dp_def}
\end{align}
where the probability is taken over the randomness of the algorithm $\mathcal{A}$, with $\phi \sim \Phi$.

% \textbf{Memorization} of the $i^{th}$ element $z_i = (x_i, y_i)$ of the dataset $S$ by an algorithm $\mathcal{A}$ was defined by \citet{feldman2019does} using the notion of stability as:
% \begin{align}
%     \mathrm{mem}(\mathcal{A}, S, i) &= \Pr_{\phi}[h_{S}^\phi(x_i) = y_i] - \Pr_{\phi}[h_{S^{\setminus i}}^\phi(x_i) = y_i],
% \label{eq:mem}
% \vspace{-4mm}
% \end{align}
% where the probability is taken over the randomness of algorithm $\mathcal{A}$.

\textbf{Error Stability}  of a possibly randomized algorithm $\mathcal{A}$ for some $\beta > 0$ is defined as \citet{kearns1997algorithmic}: %is  $\beta$ if
\begin{align}
     \forall i \in \{1, \cdots,  m\},~\left\lvert \Ex_{\phi, z}[\ell(h_S^\phi, z)] - \Ex_{\phi, z}[\ell(h^{\phi}_{S^{\setminus i}}, z)] \right\rvert~ \leq \beta,
\label{as:stability}
\end{align}
where $z \sim \mathcal{D}$ and $\phi \sim \Phi$.

\textbf{Generalization.} A randomized algorithm $\mathcal{A}$ is considered to generalize with confidence $\delta$ and at a rate of $\gamma'(m)$ if:
\begin{align}
\Pr[\left\lvert R_{emp}(h, S) - R(h) \right\rvert \leq \gamma'(m)] \geq \delta.
\label{as:gen}
\end{align}

\textbf{Uniform Model Bias.} The hypothesis $h$ produced by algorithm $\mathcal{A}$ to learn the true conditional $h^* = \Ex[y| x]$ from a dataset $S \sim \mathcal{D}^m$ has uniform bound on model bias denoted by $\Delta$ if:
\begin{align}
    \forall S \sim \mathcal{D}^m,  \quad \left\lvert \Ex_\phi[R(h_S^\phi) - R(h^*)] \right \rvert \leq \Delta.
\label{as:model_bias}
\end{align}

\textbf{$\rho$-Lipschitz Hessian.} The Hessian of $\ell$ is Lipschitz continuous on $\mathcal{Z}$, if $\forall z_1, z_2 \in \mathcal{Z}$, and $\forall h \in \mathrm{Range}(\mathcal{A})$, there exists some $\rho > 0$ such that:
\begin{align}
\lVert \nabla^2_{z_1} \ell(h, z_1) - \nabla^2_{z_2} \ell(h, z_2) \rVert \leq \rho\lVert   z_1 - z_2 \rVert.
\label{as:hess_lip}
\end{align}

\textbf{Input Loss Curvature.} As defined by \citet{moosavi2019robustness, garg2023memorization}, input loss curvature is the sum of the eigenvalues of the Hessian $H$ of the loss with respect to input $z_i$. This is conveniently expressed using the trace as:
\begin{align}
    \Curv_{\phi}(z_i, S) = \mathrm{tr}(H) = \mathrm{tr}(\nabla^2_{z_i} \ell(h^{\phi}_S, z_i))
    \label{eq:def_curv}
\end{align}

\textbf{$\upsilon$-adjacency.} A dataset $S$ is said to contain $\upsilon$-adjacent (read as upsilon-adjacent) elements if it contains two elements $z_i, z_j$ such that $z_j = z_i + \alpha$ for some $\alpha \in B_p(\upsilon)$ (read as $\upsilon$-Ball). This condition can be ensured through construction. Consider a dataset $S'$ which has no $z_j$ s.t $z_j = z_i + \alpha; z_j, z_i \in S'$. We can construct $S$  such that $S = \set{z ; z \in S'} \cup \set{z_i + \alpha}$ for some $z_i \in S', \alpha \in B_p(\upsilon)$, thereby ensuring $\upsilon$-adjacency.

\textbf{Membership Inference Threat Model.} In a membership inference security game \citep{yeom2018privacy, jayaraman2021revisiting, carlini2022membership}, a challenger and an adversary interact to test the privacy of a machine learning model. The process begins with the challenger sampling a training dataset from a distribution and training a model on this data. The challenger then flips a coin to decide whether to select a fresh data point from the distribution, which is not part of the training set, or to choose a data point from the training set. This selected data point is given to the adversary, who has access to the same data distribution and the trained model. The adversary's task is to determine whether the given data point was part of the training set or not. If the adversary's guess is correct, the game indicates a successful membership inference attack. Such a game is said to be in a \textbf{black-box} setting when the adversary has access to only the challenger's output.

\vspace{-2mm}
\section{Theoretical Analysis}
\label{sec:theory}
In this section, we analyze the distinguishability of train-test samples using KL divergence. We do so for two cases, the first case when using network's output probability, and second when using input loss curvature.
The importance of analyzing network output probabilities stems from its utilization in the current state-of-the-art attack, LiRA \citep{carlini2022membership}.
% and in prior attacks.
Thus, studying both cases will let us theoretically analyze and compare their performance with input loss curvature.

\begin{figure}[t!]
    \centering
    \begin{minipage}[t]{0.48\textwidth}
        \centering
        \includegraphics[width=\textwidth]{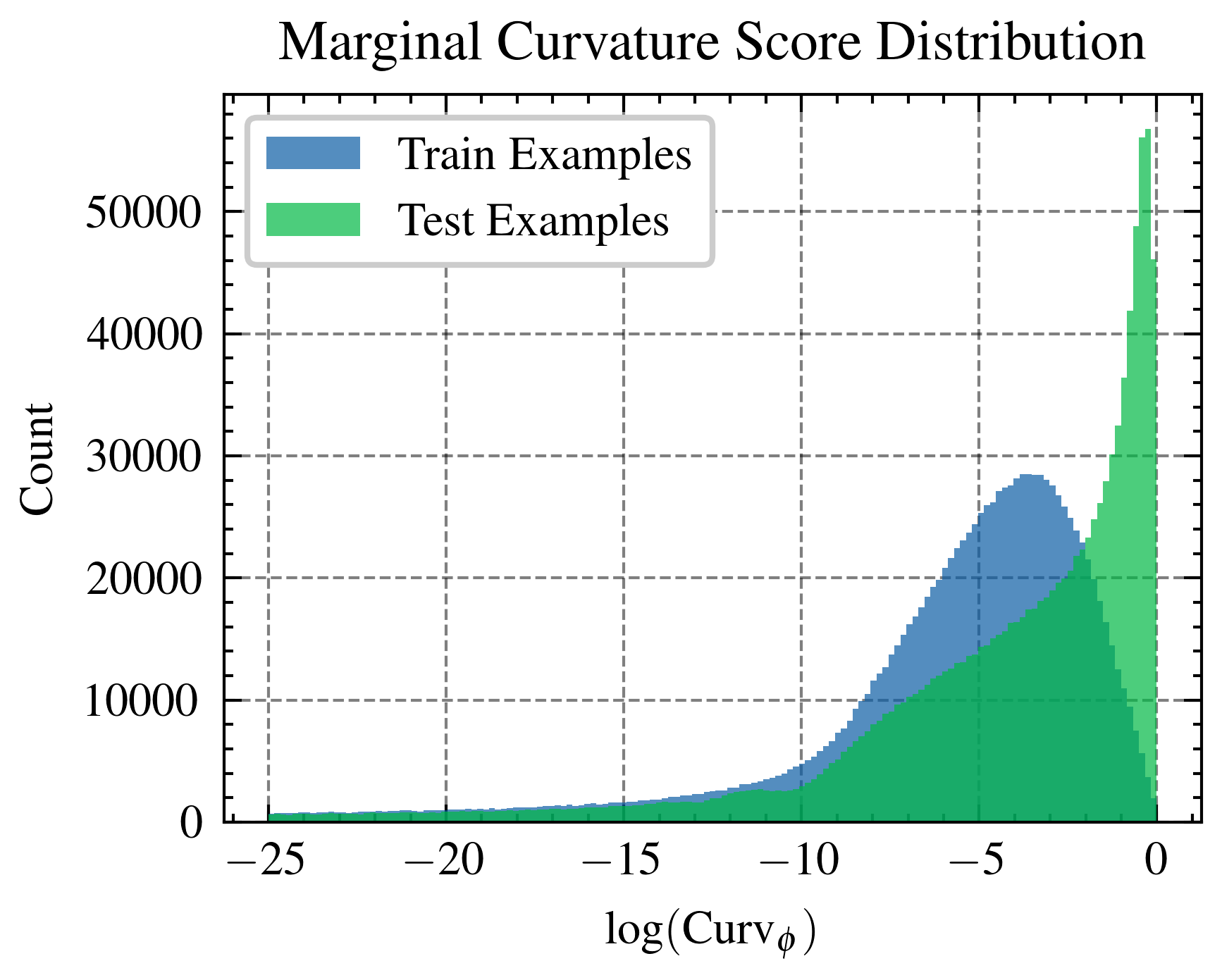}
        \vspace{-3.5mm}
        \caption{Marginal curvature score histogram for all images in ImageNet when samples in train set vs test set.}
        \label{fig:log_curv_marginal}
    \end{minipage}\hfill
    \begin{minipage}[t]{0.48\textwidth}
        \centering
        \includegraphics[width=0.96\textwidth]{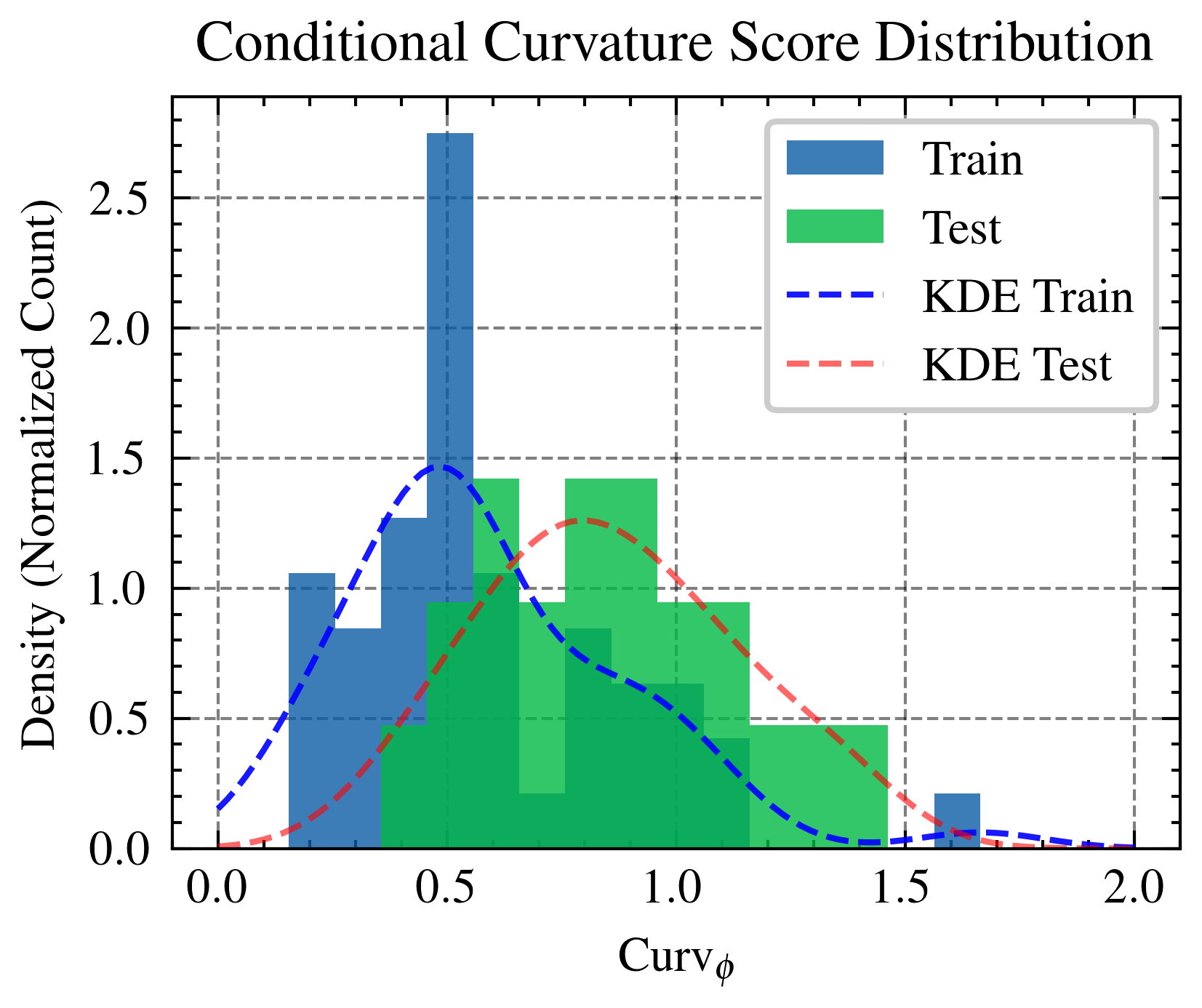} % Second image
        \caption{Conditional curvature scores for a single image from ImageNet when using 56 shadow models.}
        \label{fig:curv_conditional}
    \end{minipage}
    \vspace{-2mm}
\end{figure}

Before we begin the analysis, we briefly discuss the conditional and marginal distributions of input loss curvature for train and test examples. Discussing this is important to build intuition for the analysis. Figure \ref{fig:log_curv_marginal} visualizes the histogram (proxy for distribution) of input loss curvature for train and test examples from ImageNet \citep{ILSVRC15}. Specifically, it plots the log of the input loss curvature $\log(\Curv_{\phi})$ obtained on pre-trained ResNet50 \citep{he2016deep} models from \citet{feldman2020neural}. It plots a histogram of these scores, showing the frequency of specific log curvature values. 

A naive membership inference attack would apply a threshold to separate these distributions. However, it is common practice to consider sample specific scores (i.e., conditioned on the target sample). This is visualized in Figure \ref{fig:curv_conditional} which plots the distribution of curvature scores for a single ImageNet sample, indicating differences when the sample is part of the training set versus when it is a test or unseen sample. The data was generated using models from \citet{feldman2020neural}. The figure also includes a kernel density estimate (KDE, shown as a dashed line) to better to visualize the underlying distribution. Figure \ref{fig:curv_conditional} suggests that, similar to \citet{carlini2022membership} sample conditional curvature scores can be modeled using a Gaussian parametric model. If we represent the curvature score $\Curv_{\phi}$ as a random variable, then     $\Curv_{\phi} \sim \mathcal{N}(\mu, \sigma)$. The probability density function of $\Curv_{\phi}$ is a function of the randomness of the algorithm $\phi$, the dataset $S$ and the $i^{th}$ sample $z_i$ and be denoted by $p_c(\phi, S, z_i)$. Similarly, let $p(\phi, S, z_i)$ denote the probability density function of the neural network's output probability, which is also parameterized by the randomness of the algorithm $\phi$, the dataset $S$ and the $i^{th}$ sample $z_i$. With this setup we present the following theoretical results on the upper bound on the KL divergence between train and test distribution for the two cases. Theorem \ref{th:p_dkl} presents the upper bound on the KL divergence when using the neural network's output probability scores, and Theorem \ref{th:pc_dkl} presents the upper bound on the KL divergence when using input loss curvature.

\begin{theorem}[Privacy bounds Train-Test KL Divergence]  Assume $\epsilon$-differential
private algorithm, then the KL divergence between train-test distributions of the neural network's output probability is upper bound by the differential privacy parameter $\epsilon$ given by:
\label{th:p_dkl}
\begin{align}
\KL\left( p(\pmb{\phi}, S, z_i)~||~p(\pmb{\phi}, S^{\setminus i}, z_i) \right) &\leq \epsilon
\end{align}
\end{theorem}

% Next we present Lemma \ref{lm:p_dkl_curv}, which provides a link between the KL divergence of train-test distributions of the neural network's output and input loss curvature.  

% \begin{lemma}[Test Curvature bounds Train-Test KL Divergence] For any randomized algorithm which is $\epsilon$-differentially private and the assumptions of error stability \ref{as:stability}, generalization \ref{as:gen}, and uniform model bias \ref{as:model_bias} hold. Further, assume $0 \leq \ell \leq L$. Then for two adjacent datasets $S, S^{\setminus i} \sim \mathcal{D}$ with a probability at least $1-\delta$ we have 
% \label{lm:p_dkl_curv}
% \begin{align}
%     \KL\left( p(\pmb{\phi}, S, z_i)~||~p(\pmb{\phi}, S^{\setminus i}, z_i) \right) \leq \Ex_{\pmb{\phi}}[\Curv(z_i, S^{\setminus i})]  + c_1 \\
%     c_1 = m\beta + (4m-1)\gamma + 2(m-1)\Delta + \cfrac{\rho}{6} \Ex[\lVert \alpha \rVert^3] 
% \end{align}
% \end{lemma}
\begin{theorem}[Dataset Size and Privacy bound Curvature KL Divergence]  Let the assumptions of error stability \ref{as:stability}, generalization \ref{as:gen}, and uniform model bias \ref{as:model_bias} hold. Further, assume $0 \leq \ell \leq L$. Let the conditional distribution be parameterized with variance $\sigma$. Then, any $\epsilon$-differential private algorithm using a dataset of size $m$ with a probability at least $1-\delta$ satisfies:
\label{th:pc_dkl}
\begin{align}
\KL(p_c(\pmb{\phi}, S, z_i)~||~p_c(\pmb{\phi}, S^{\setminus i}, z_i)) &\leq \cfrac{\left[Lm(1- e^{-\epsilon}) + c \right]^2}{2 \sigma^2} \label{eq:th_pc_dkl} \\
c = (4m-1)\gamma +  2 (m-1)&\Delta  +  \cfrac{\rho}{6} \Ex[\lVert \alpha \rVert^3] + L
\end{align}
\end{theorem}

\textbf{Discussion.} The proofs for Theorems \ref{th:p_dkl} and \ref{th:pc_dkl} are provided in Appendix \ref{appndx:th_p_dkl} and \ref{appndx:th_pc_dkl} respectively. Theorem \ref{th:p_dkl} and Theorem \ref{th:pc_dkl} provide the upper bound on KL Divergence between train and test distributions of the probability and input loss curvature scores, respectively. They imply the upper limit of MIA performance, and interestingly, the result of Theorem \ref{th:pc_dkl} suggests the role of the dataset size on MIA performance. However, the relation is not as straightforward as suggested by Equation \ref{eq:th_pc_dkl}. This is because, in real applications, parameter $\sigma$ depends on $\epsilon$, and so does the loss bound $L$. In fact, the loss bound is also dependent on the number of samples $m$ \citep{bousquet2002stability}.
% Thus for convenience we modelled it as $\zeta(\epsilon, m) = L(m, \epsilon) / 2\sigma(\epsilon)^2$.

\begin{theorem}[Dataset Size and Curvature MIA Performance] 
\label{th:m_limit}
Let the assumptions of error stability \ref{as:stability}, generalization \ref{as:gen}, and uniform model bias \ref{as:model_bias} hold. Further, assume $0 \leq \ell \leq L$, and the bounds of Theorem \ref{th:p_dkl} and \ref{th:pc_dkl}  are tight. Then, the performance of MIA using curvature scores exceeds that of confidence scores with a probability at least $1-\delta$ when:
\begin{align}
    % L(m+1)(1-e^\epsilon) &\geq \epsilon\\
    % \implies 
    m &>  \cfrac{(\sqrt{2\sigma^2\epsilon}) - c }{L(1-e^{-\epsilon})} %- 1
\end{align}
\end{theorem}
Theorem \ref{th:m_limit} suggests that the performance of curvature based MIA will exceed that of probability score based methods when the size of the dataset used to train the target model exceed a certain threshold. Indeed, this is what we observe in practice (see section \ref{sec:effect_of_m}).

\textbf{On the validity of the assumptions.} The proof for Theorem \ref{th:m_limit} can be found in Appendix \ref{appndx:th_m_limit}. Before presenting our experiments to validate our theory, we briefly discuss the validity of our assumptions in practical settings. Research by \citet{hardt2016train} shows that stochastic gradient methods, such as stochastic gradient descent, achieve small generalization error and exhibit uniform stability. Thus, the assumptions of stability (Equation \ref{as:stability}) and generalization (Equation \ref{as:gen}) are justified.  Model bias is a property of the model, and a uniform bound across different datasets seems reasonable. Note, uniform loss bound (and $L$ independent of $m$ as used by Theorem \ref{th:m_limit}) holds true for certain losses and for statistical models and is often assumed in learning theory \citep{wang2016learning}. Lastly, the $\upsilon$-adjacency can be ensured through construction. For large datasets this may not be needed, because the size of the ball $B_p(\upsilon)$ is unconstrained. Hence two samples from the same class that are similar may suffice. Given the size of modern datasets, this assumption is also reasonable.
% \vspace{-4mm}

\section{Zero Order Input Loss Curvature MIA}
To test if input loss curvature based membership inference performs better than existing methods we need an efficient technique to estimate curvature. We are interested in black-box membership inference attacks, where one does not have access to the target network's parameters. However, current techniques use Hutchinson's trace estimator \citep{hutchinson1989stochastic} to measure input loss curvature such as from \citet{garg2023samples}, \citet{garg2023memorization} or \citet{ravikumar2024unveiling}. 
% given by:
% \begin{align}
% % \quad \mathrm{tr}(H^2) &\propto \frac{1}{n} \sum_{i=0}^{n} \left\lVert \frac{\partial  \left(L(x+hv) - L(x)\right)}{\partial x} \right\rVert^2_2 \nonumber \\
% \Curv(z_i) &\propto \frac{1}{n} \sum_{i=0}^{n} \left\lVert \frac{\partial  \left(\ell(z_i+hv) - \ell(z_i)\right)}{\partial z_i} \right\rVert^2_2
%     \label{eq:curv}
% \end{align}
% where $n$ is the number of Rademacher vectors to average and $h$ is a hyper parameter.
This approach needs to evaluate the gradient and hence requires access to model parameters. To solve this issue, we propose using a zero-order estimation technique.
Zero-order curvature estimation starts with a finite-difference estimation. Consider a function \( f: \mathbb{R}^n \rightarrow \mathbb{R} \), then the Hessian at a given point \( z_i \) can be estimated as follows:
\begin{align}
    \nabla^2 f(z_i) = n^2 \frac{ f(z_i + h v + h u) - f(z_i - h v + h u) - f(z_i + h v - h u) + f(z_i - h v - h u)}{4 h^2}uv^\top
    \label{eq:zo_base}
\end{align}

where $h$ is a small increment (a hyper parameter in out case), and \( u, v \) are vectors in \( \mathbb{R}^n \). In our case, to get the input loss curvature, we have  $f \gets \ell(g(x_i), y_i)$, where $g$ is the neural network, $\ell$ is the loss function and $z_i = (x_i, y_i)$ are the image, label pair. The pseudo-code for obtaining input loss curvature score  using zero order estimation shown in Algorithm \ref{alg:pseudo-zo-curv} in Appendix \ref{sec:appndx_zo_alg}. To execute membership inference attack using input loss curvature scores, we propose the following methodology. First, we begin by training shadow models, similar to \citet{shokri2017membership, carlini2022membership}. These shadow models are used to obtain empirical estimates of parametric model for the curvature score described in Section \ref{sec:theory}. During the inference phase, we employ a likelihood ratio between the sample being in the train set vs test set parametric models to identify the membership status of a given sample (see Appendix \ref{appndx:mia_curvature} for pseudo-code). In addition, we perform a negative log likelihood ratio test. We denote the results of likelihood test as `LR' and the results of the negative log likelihood ratio test as `NLL'.

\section{Experiments}
% In this section, we perform experiments to validate the links predicted by theory. 
\subsection{Experimental Setup}
\label{sec:exp_setup}
\textbf{Datasets.} To evaluate our theory, we consider the classification task using standard vision datasets. Specifically, we use the CIFAR10, CIFAR100 \citep{krizhevsky2009learning} and ImageNet \citep{ILSVRC15} datasets. 

\textbf{Architectures.} For experiments on ImageNet we use 
% the models released by \citet{feldman2020neural} which used
the ResNet50 architecture \citep{he2016deep}. For CIFAR10 and CIFAR100, we used the ResNet18 architecture. Details regrading hyperparameters are provided in Appendix \ref{appndx:reprod_docu}. To improve reproducibility, we have provided the code in the supplementary material.
% Details regarding the model used are specified at the beginning of each experiment section. 
% Details regrading hyperparameters, training and testing  are provided in Appendix \ref{sec:implementation}. To improve reproducibility, we have provided the code in the supplementary material.

\textbf{Training.} For experiments using private models, we trained ResNet18 models with the Opacus library \citep{opacus}  using DP-SGD with a maximum gradient norm of 1.0 and a privacy parameter of $\delta=1\times 10^{-5}$. 
% The initial learning rate was 0.001, decreased at epochs 12 and 16, with a batch size of 128.
Shadow models for CIFAR10 and CIFAR100 were trained on a 50\% subset of the data for 300 epochs.
% , using an initial learning rate of 0.1, decayed at the 180th and 240th epochs. 
For ImageNet, we used pre-trained models from \citet{feldman2020neural}, trained on a 70\% subset of ImageNet. More details about training and compute resources are provided in Appendix \ref{appndx:reprod_docu} and we have provided the code in the supplementary material.

% \textbf{Testing.} 
% During testing we used resize followed by normalization. We used two augmentations the original image and its mirror. The number of augmentations used are specified in the corresponding experiment section. When using pre-trained models from \citet{feldman2020neural} we validated the accuracy of the models before performing experiments. To improve reproducibility, we have provided the code in the supplementary material. All of the experiments were performed on a heterogeneous compute cluster consisting of 9 1080Ti's, 6 2080Ti's and 4 A40 NVIDIA GPUs, with a total of 100 CPU cores and a combined 1.2 TB of main system memory. However, the results can be replicated with a single GPU with 11GB of VRAM. 

\textbf{Metrics.} To evaluate curvature scores, we use AUROC and balanced accuracy. The Receiver Operating Characteristic (ROC) is the plot of the True Positive Rate (TPR) against the False Positive Rate (FPR). The area under the ROC is called Area Under the Receiver Operating Characteristic (AUROC). AUROC of 1 denotes an ideal detection scheme, since the ideal detection algorithm results in 0 false positive and false negative samples.  
% Recall (TPR) and FPR are defined by equations (\ref{eqn:TPR}) and (\ref{eqn:FPR}) respectively. 
 
% \begin{equation}\label{eqn:TPR}
%     \begin{array}{cc}
%       \text{Recall = True Positive Rate (TPR)} &= \cfrac{TP}{TP+FN}
%     \end{array}
% \end{equation}
% \begin{equation}\label{eqn:FPR}
%     \begin{array}{cc}
%          \text{False Positive Rate (FPR)} &= \cfrac{FP}{FP+TN}
%     \end{array}
% \end{equation}
% where, TP is True Positive, FP is False Positive, TN is True Negative and FN is False Negative. 

% For all our experiments we used $h=1\times10^{-3}$ and $n=10$. We found the results to be robust to changes in $h$; we varied it from $1\times10^{-1}$ to $1\times10^{-3}$. We also varied $n$ from $5,10,20$ and found the results to be robust to changes in $n$.% 

\subsection{Membership Inference}

In this subsection, we compare the performance of the proposed input loss curvature based membership inference against prior MIA techniques.

\begin{figure}[t!]
    \centering
    \begin{minipage}[t]{0.48\textwidth}
        \centering
        \includegraphics[width=\textwidth]{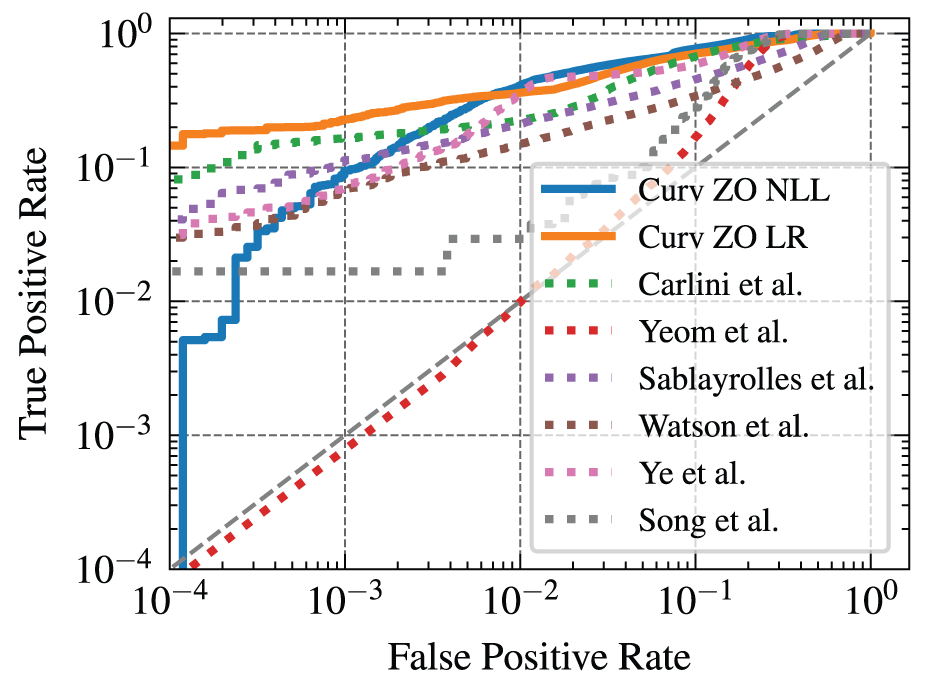}
        \vspace{-4mm}
        \caption{Comparing our method against existing techniques at \emph{low FPR}. 
        % We used CIFAR100 dataset with 64 shadow models. 
        The proposed parametric Curv LR technique has the highest TPR at very low FPR.}
        \label{fig:cifar100}
    \end{minipage}\hfill
    \begin{minipage}[t]{0.475\textwidth}
        \centering
        \includegraphics[width=\textwidth]{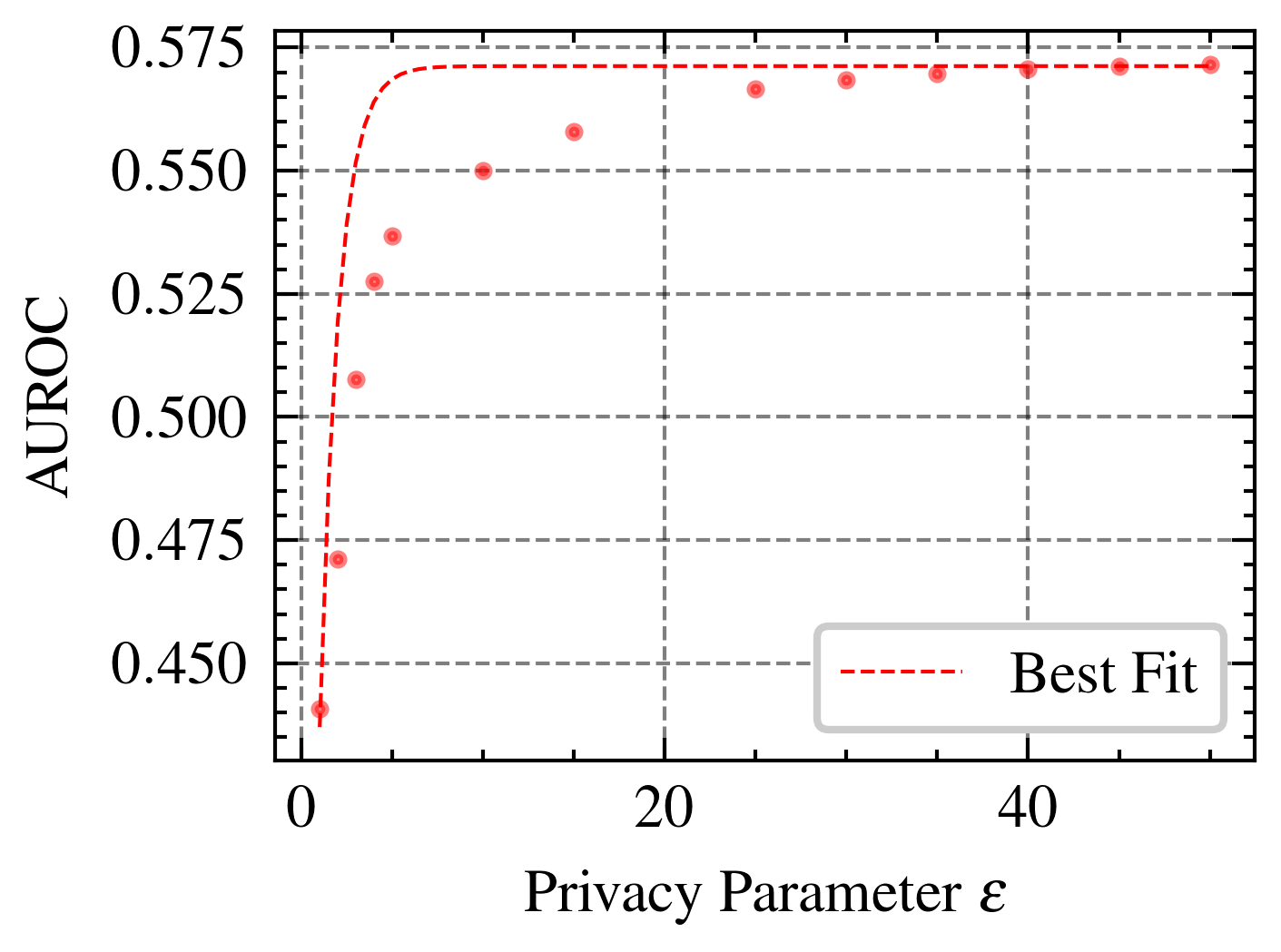}
        \vspace{-4mm}
        \caption{Validating the upper bound from Theorem \ref{th:pc_dkl} by fitting the MIA performance (AUROC of Curv LR) on DP-SGD trained models for various privacy parameters $\epsilon$ values.}
        \label{fig:eps_vs_mia}
    \end{minipage}
    \vspace{-4mm}
\end{figure}

\textbf{Setup:} We use CIFAR10, CIFAR100 and ImageNet datasets to test the MIA performance. We consider a \textbf{black-box} MIA setup similar to \citet{carlini2022membership}. We use ResNet18 for CIFAR10 and CIFAR100, for ImageNet we use the ResNet50 architecture. For all the membership inference attacks, we compute a full ROC curve and report the results. When using shadow models we 64 for CIFAR10 and CIFAR100 and 52 for ImageNet. The AUROC plot for CIFAR100 for various methods are shown in Figure \ref{fig:cifar100}. Table \ref{tab:mia_results} reports the average balanced accuracy and AUROC values over three seeds on various MIA methods on the three datasets.
The plot in Figure \ref{fig:cifar100} is a log-log plot to emphasize performance of the proposed method at very low false positive rates (see the orange line y-intercept).
We also studied the effect of augmentations and the results can be found in Appendix \ref{appndx:aug_results}, the takeaway was that adding more augmentations improved performance. Note that the results presented in Table \ref{tab:mia_results} used 2 augmentations (image + mirror) for our technique and \citet{carlini2022membership} for fair comparison.
% \textcolor{red}{ablation on number of augs, add std to table 1, add tpr 0.1\% results}

\textbf{Results:} From Table \ref{tab:mia_results}, we see that the proposed method performs better than all existing MIA techniques on both ImageNet and CIFAR datasets. Apart from AUROC and balanced accuracy results, the log-log plot emphasizes the performance at really low FPR (i.e. the y intercept, high TPR at low FPR in Figure \ref{fig:cifar100}). Additional results at low FPR are available in Appendix \ref{appndx:tpr_fpr}.

\textbf{Takeaways:} 
% We could not not scale \citet{jayaraman2021revisiting} to ImageNet due to compute limitations as the method proposed by \citet{jayaraman2021revisiting} uses 100 augmentations per image. Thus is left blank.
As predicted by Theorem \ref{th:pc_dkl}, the higher KL divergence between train and test curvature score distributions results in superior MIA performance. Further, we observe that while using a negative log likelihood ratio test does better in AUROC and balanced accuracy, the parametric likelihood ratio test does better at low false positive rates as shown in Figure \ref{fig:cifar100}. Thus, the proposed use of curvature should be tailored based on use case. Applications that demand high AUROC can use NLL approach, while those that demand high TPR at very low FPR should use the LR technique.
% We found that the negative log likelihood ratio test is more sensitive to the number of shadow model used.
\vspace{-2mm}
\begin{table}[b!]
\renewcommand*{\arraystretch}{1.2}
\begin{adjustbox}{width=\textwidth}
\centering
\begin{tabular}{lcccccccc}
\toprule
\multirow{2}{*}{Method} & \multicolumn{2}{c}{ImageNet} & & \multicolumn{2}{c}{CIFAR100} & & \multicolumn{2}{c}{CIFAR10} \\
\cmidrule{2-3} \cmidrule{5-6} \cmidrule{8-9}
 & Bal. Acc. & AUROC & & Bal. Acc. & AUROC & & Bal. Acc. & AUROC \\
\midrule
\textbf{Curv ZO NLL (Ours)} & \textbf{69.16 {\scriptsize$\pm$ 0.08}} & \textbf{77.45 {\scriptsize$\pm$ 0.09}} & & \textbf{84.47 {\scriptsize$\pm$ 0.21}} & \textbf{93.49 {\scriptsize$\pm$ 0.18}} & & \textbf{61.92 {\scriptsize$\pm$ 0.87}} & \textbf{68.82 {\scriptsize$\pm$ 1.30}} \\
Curv ZO LR (Ours) & 68.76 {\scriptsize$\pm$ 0.04} & 72.28 {\scriptsize$\pm$ 0.04} & & 80.48 {\scriptsize$\pm$ 0.10} & 90.15 {\scriptsize$\pm$ 0.04} & & 55.00 {\scriptsize$\pm$ 0.17} & 58.89 {\scriptsize$\pm$ 0.38} \\
% Curv ZO 1-NLL (Ours) & 64.90 {\scriptsize$\pm$ 0.10} & 73.15 {\scriptsize$\pm$ 0.10} & & - & - & & - & - \\
\cite{carlini2022membership} & 66.14 {\scriptsize$\pm$ 0.01} & 73.46 {\scriptsize$\pm$ 0.02} & & 81.55 {\scriptsize$\pm$ 0.13} & 88.89 {\scriptsize$\pm$ 0.16} & & 58.23 {\scriptsize$\pm$ 0.29} & 61.73 {\scriptsize$\pm$ 0.32} \\
\cite{yeom2018privacy} & 58.50 {\scriptsize$\pm$ 0.02} & 63.23 {\scriptsize$\pm$ 0.03} & & 76.29 {\scriptsize$\pm$ 0.39} & 82.11 {\scriptsize$\pm$ 0.31} & & 55.57 {\scriptsize$\pm$ 0.52} & 60.44 {\scriptsize$\pm$ 0.75} \\
\cite{sablayrolles2019white} & 66.93 {\scriptsize$\pm$ 0.05} & 76.50 {\scriptsize$\pm$ 0.04} & & 70.22 {\scriptsize$\pm$ 0.41} & 81.11 {\scriptsize$\pm$ 0.39} & & 56.65 {\scriptsize$\pm$ 0.56} & 61.50 {\scriptsize$\pm$ 0.79} \\
\cite{watson2021importance} & 61.40 {\scriptsize$\pm$ 0.06} & 69.44 {\scriptsize$\pm$ 0.05} & & 62.71 {\scriptsize$\pm$ 0.31} & 71.66 {\scriptsize$\pm$ 0.50} & & 54.86 {\scriptsize$\pm$ 0.59} & 58.58 {\scriptsize$\pm$ 0.86} \\
\cite{ye2022enhanced} & 66.16 {\scriptsize$\pm$ 0.02} & 75.79 {\scriptsize$\pm$ 0.05} & & 80.73 {\scriptsize$\pm$ 0.24} & 90.88 {\scriptsize$\pm$ 0.19} & & 59.62 {\scriptsize$\pm$ 0.84} & 67.30 {\scriptsize$\pm$ 1.25} \\
\cite{song2021systematic} & 57.88 {\scriptsize$\pm$ 0.03} & 63.29 {\scriptsize$\pm$ 0.03} & & 75.58 {\scriptsize$\pm$ 0.29} & 82.28 {\scriptsize$\pm$ 0.27} & & 55.63 {\scriptsize$\pm$ 0.61} & 60.42 {\scriptsize$\pm$ 0.85} \\
% \cite{jayaraman2021revisiting} & - & - & & 70.33 & 69.05 & & 60.14 & 62.51 \\
\bottomrule
\end{tabular}
\end{adjustbox}
\vspace{2mm}
\caption{Comparison of the proposed curvature score based MIA with prior methods tested on ImageNet, CIFAR100, and CIFAR10 datasets. Results reported are the mean $\pm$ std obtained over 3 seeds. For CIFAR10 and CIFAR100 64 shadow models were used and 52 for ImagNet.}
\label{tab:mia_results}
\end{table}

\subsection{Effect of Privacy}
In this section, we study the effect of differential privacy on MIA performance and test the $\epsilon$ relation predicted by Theorem \ref{th:pc_dkl}. 
% which predicts performance  is bound by $\propto (1-e^{-\epsilon})$.

\textbf{Setup}: To study the effect of privacy on MIA attack performance, we use DP-SGD trained models. We use privacy $\epsilon$ values of 1, 2, 3, 4, 5, 10, 15, 20, 25, 30, 35, 40, 45, 50 with $\delta = 1 \times 10^{-5}$. We plot the result of the proposed curvature based MIA in Figure \ref{fig:eps_vs_mia}. In Figure \ref{fig:eps_vs_mia} we also plots the best fit using $s_f (L_f  (1-e^{-\epsilon}) + c_f )^2$, where $s_{f}, L_f$ and $ c_f$ are fit to the data. 

% \textcolor{red}{talk why it matches well}

\textbf{Results and Takeaways:} We see that the theoretical prediction from Theorem \ref{th:pc_dkl} about MIA performance is well matched. Since Theorem \ref{th:pc_dkl} provides an upper bound, the results validate the theory.
% Further, we see that a lower $\epsilon$ i.e. higher privacy LiRA \citep{carlini2022membership} performs better (i.e. upto $\approx$ 7) after which the propose curvature based MIA performs better.

\subsection{Effect of Dataset Size}
\label{sec:effect_of_m}
In this section, we study the effect of dataset size on MIA attack performance. This lets us test the relationship of MIA performance to $m$ as predicted by Theorem \ref{th:pc_dkl} and Theorem \ref{th:m_limit}. 

\begin{figure}[t!]
    \centering
    \begin{minipage}[t]{0.48\textwidth}
        \centering
        \includegraphics[width=\textwidth]{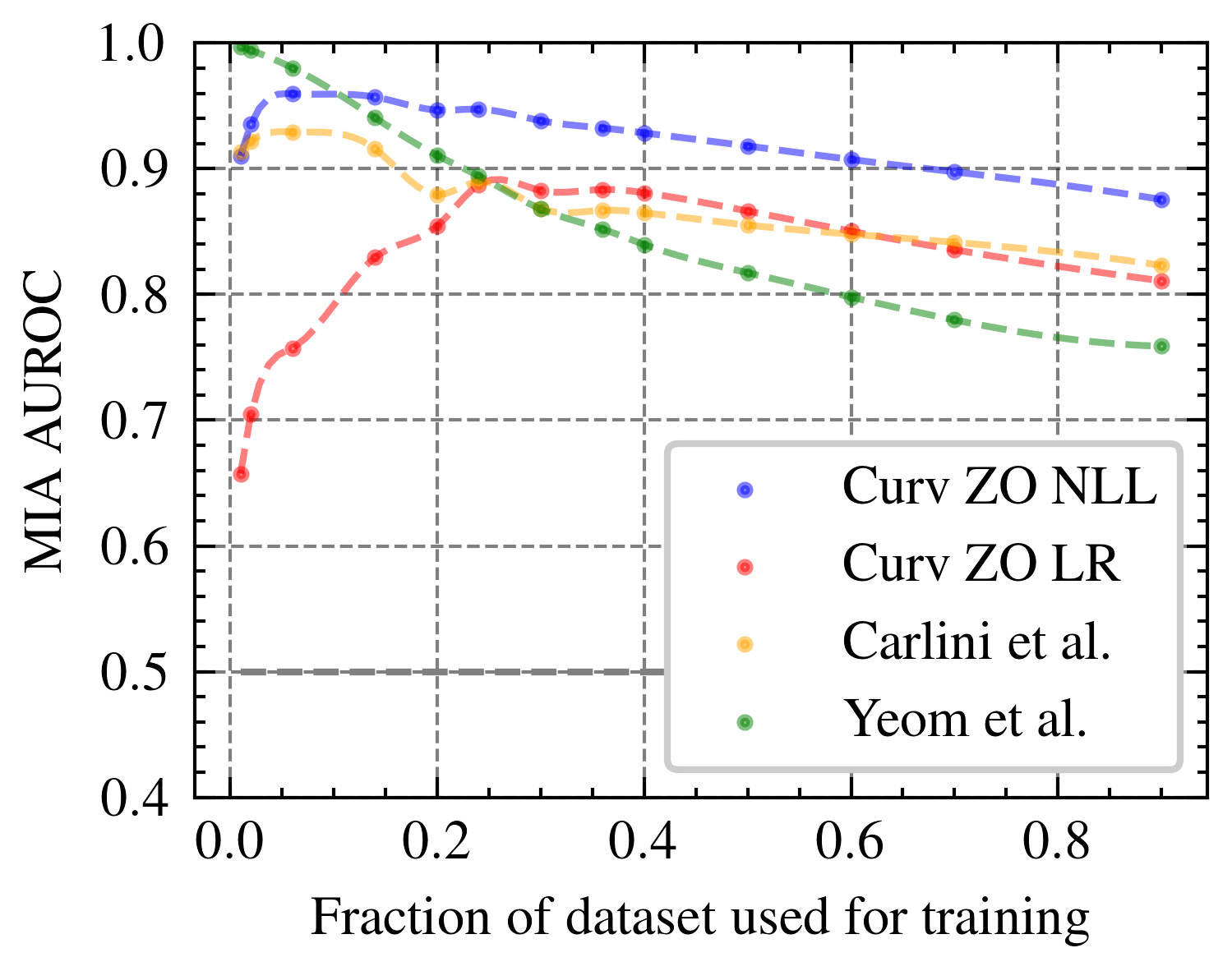}
        \vspace{-4mm}
        \caption{Visualizing MIA performance as a function of the size of the train set, which is randomly sampled.}
        \label{fig:m_v_random}
    \end{minipage}\hfill
    \begin{minipage}[t]{0.49\textwidth}
        \centering
        \includegraphics[width=\textwidth]{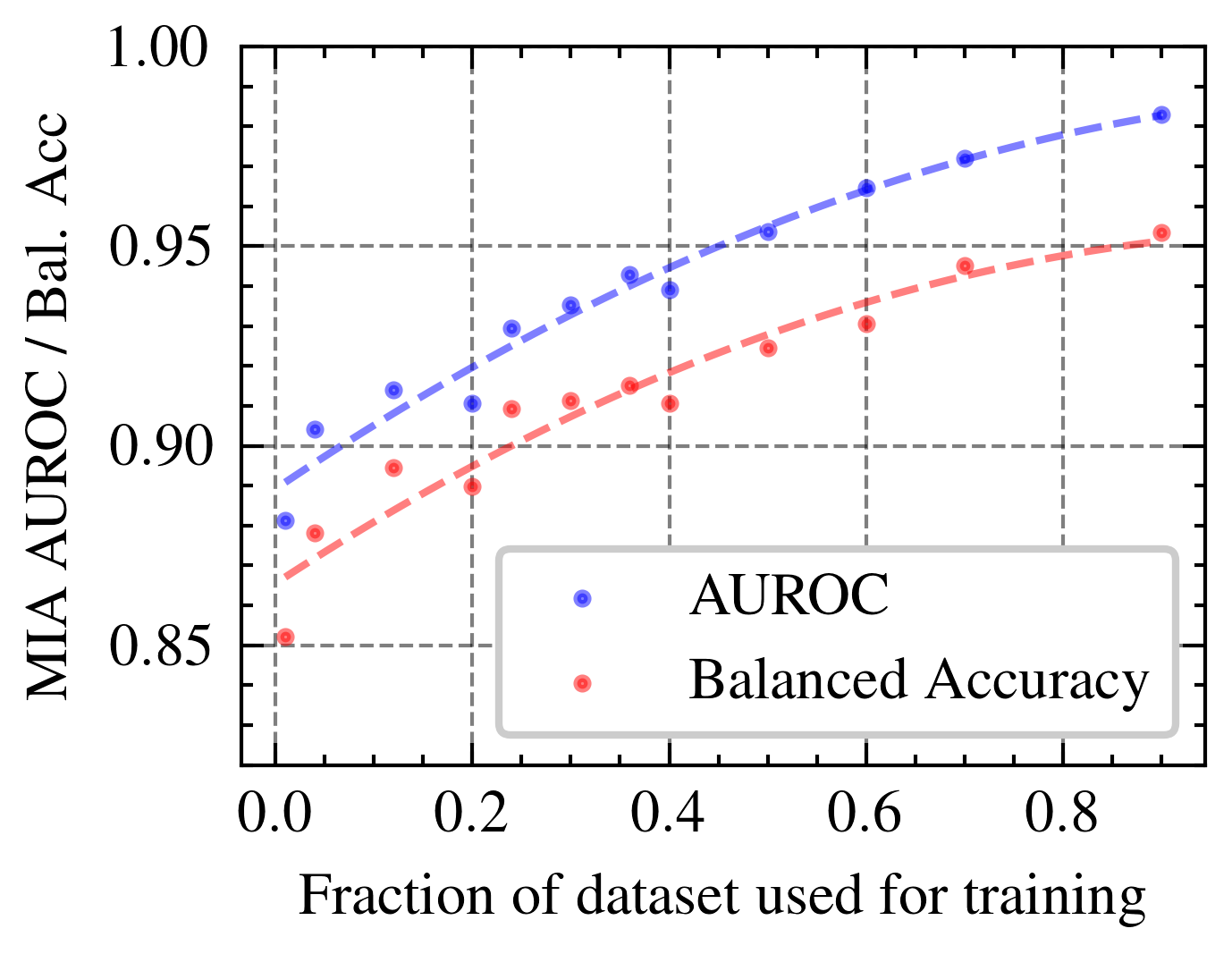}
        \vspace{-4mm}
        \caption{MIA performance as a function of the size of the train set when subsets contain the lowest curvature examples from the entire set.}
        \label{fig:m_v_top}
    \end{minipage}
    % \vspace{-4mm}
\end{figure}

\textbf{Setup:} For this experiment, we train models on increasing dataset size on CIFAR100. Specifically, we train multiple seeds on various subsets randomly chosen from the CIFAR100 training set. We also repeated the same by choosing the lowest curvature samples from CIFAR100 and train with the same subset sizes. Next, for each of the models we perform MIA attack and we present the results in Figure \ref{fig:m_v_random} (randomly chosen) and Figure \ref{fig:m_v_top} (lowest curvature samples).

\textbf{Results:} From Figure \ref{fig:m_v_random}, we see that when samples are randomly chosen, the MIA performance decreases as we add more samples. This result is consistent with prior literature \citep{abadi2016deep}. Further as predicted by Theorem \ref{th:m_limit} beyond $30-40\%$ subset Curv ZO LR out perform prior works and Curv ZO NLL outperforms prior works beyond $10\%$ subset size.  
% up to a point and then increases. 
% This is because initially small subsets are memorized by the network, as we add more samples the network generalizes more, which reduces MIA performance and finally, as even more samples are added the network memorizes again. 
% This is because the shadow models were trained on 50\% subset. This is clear from the fact that \citet{yeom2018privacy} which is a shadow model free technique and does not show this trend. \textbf{This reveals a previously unknown limitation of shadow model based methods.}

To extend this analysis, we train models on curvature based coresets \citep{garg2023samples}. These coresets of low input loss curvature samples have been shown to memorize less \citep{garg2023memorization, ravikumar2024unveiling}. Thus we expect MIA accuracy to increase as we increase coreset size. This is exactly what happens and is shown in Figure \ref{fig:m_v_top} which plots the AUROC and accuracy of the NLL curvature attack as we increase curvature coreset size. However, the MIA performance is higher for comparable size in Figure \ref{fig:m_v_random} and \ref{fig:m_v_top}. \textbf{This suggests that curvature based coresets result in more susceptible models}, which is also supported by results in \citet{song2019privacy}.

% The motivation for choosing the lowest curvature samples was that, low curvature samples have more support in the train set thus are likely to be more private. 

\textbf{Takeaways:} 
We validate Theorem \ref{th:m_limit}. We note that beyond a certain dataset size (of about $30-40\%$ subset of the training set Curv ZO LR out perform prior works and Curv ZO NLL outperforms prior works beyond $10\%$ subset size) curvature scores outperform probability score based MIA method. While Theorem \ref{th:m_limit} predicts that curvature-based MIA outperforms other methods on sufficiently large datasets. This condition is often met by real datasets as evident from the results presented.
% We uncover a previously unknown limitation of shadow model based methods, namely shadow model-based methods perform poorly when the subset size used to train them is similar to the size of the dataset on which the target model is trained. Thus, it can be leveraged as a defense technique. Therefore, ML privacy practitioners testing privacy via MIA should consider evaluating on multiple methods and include methods without shadow models for a comprehensive test.

% Further, we see that models trained on small coresets (about 40\% $\sim$ 60\% of the dataset) are quite robust to MIA attacks. \textcolor{red}{Take away about low curvature coresets are more susptible ontraty to intution, talk about the fact that shadow models are not always the best, reduce the strength of the claim}

% \begin{figure}
%     \centering
%     \includegraphics{images/cifar100_m_v_mia.png}
%     \caption{Comparing our method against existing techniques at low FPRs. We used CIFAR100 dataset with 64 shadow models. We also report the AUROC (i.e. AUC) for each of the methods compared.}
%     \label{fig:enter-label}
% \end{figure}

% \vspace{-4mm}
\section{Conclusion}
In this paper, we explored the properties of input loss curvature on the test set. Specifically, we focus on using input loss curvature to distinguish between train and test examples. We established a theoretical framework deriving an upper bound on train-test KL Divergence based on privacy and training set size.
To extend the applicability of input loss curvature computation to a black-box setting, we propose a novel zero-order curvature estimation method. This innovation enables the development of a new cutting-edge black-box membership inference attack (MIA) methodology that leverages input loss curvature.
Through extensive experiments on the ImageNet and CIFAR datasets, we demonstrate that our input loss curvature-based MIA method outperforms existing state-of-the-art techniques. Our empirical results corroborate our theoretical predictions regarding the relationship between MIA performance and dataset size. Specifically, we show that beyond a certain threshold, the effectiveness of curvature scores surpasses other methods. Importantly, we observe that this dataset size condition is frequently met in practical scenarios, as evidenced by our results on the CIFAR100 dataset.
% To adapt curvature computation to  black-box setting we propose a zero-order curvature estimation which lead to the development of a novel black-box MIA methodology utilizing input loss curvature. Experiments on ImageNet and CIFAR datasets demonstrated that input loss curvature based MIA outperforms existing techniques. We validate our theoretical predictions regarding the relationship between MIA performance and dataset size, showing that beyond a certain dataset size, curvature scores outperform other methods. Further, we show that in practice the dataset size condition is often met by real datasets as evident from the results on CIFAR100.
% Finally, our experiments on the effect of dataset size suggest that models trained on small subsets of the training set demonstrate robustness against shadow models based MIA. This reveals a previously unknown limitation of shadow models based MIA. Hence, practitioners should consider evaluating privacy using multiple MIA methods and include techniques without shadow models for a comprehensive test. 
These findings advance our understanding of input loss curvature in the context of privacy and help build more secure deep learning models. 

\section*{Acknowledgment}

This work was supported in part by, the Center for the Co-Design of Cognitive Systems (COCOSYS), a DARPA-sponsored JUMP 2.0 center, the Semiconductor Research Corporation (SRC), and the National Science Foundation.

% \textcolor{red}{TODO checklist }
\bibliographystyle{abbrvnat}
\bibliography{egbib}

%%%%%%%%%%%%%%%%%%%%%%%%%%%%%%%%%%%%%%%%%%%%%%%%%%%%%%%%%%%%

\newpage
\appendix

\section{Appendix}

\subsection{Zero Order Curvature Estimation}
\label{sec:appndx_zo_alg}
We present the pseudo-code for zero-order curvature score calculation below. Please note that Algorithm \ref{alg:pseudo-zo-curv} shows the input loss curvature calculation for one example; however, it can be easily implemented for batch data.
\begin{algorithm}

\caption{Pseudo-code for Zero Order Input Loss Curvature Estimation}
\label{alg:pseudo-zo-curv}
\begin{algorithmic}[1] % The number tells where the line numbering should start
\State \textbf{Input:} $z_i = (x_i, y_i)$ (image, label data), $\ell$ (loss function), $n_{iter}$ (hyperparameter: number of iterations), $h$ (hyperparameter: small increment), $g$ (neural network)
\State  \textbf{Output}: $curv\_estimate$  (zero order input loss curvature estimate of $z_i$)
\State $curv \gets 0$
\For {$i \gets 1$ \textbf{to} $n_{iter}$}
    \State $v \sim \text{Rademacher}$
    \State $u \sim \text{Rademacher}$
    \State $f \gets \ell(g(x_i), y_i)$
    \State $\nabla^2 f(z_i) \gets $ \text{Use Equation \ref{eq:zo_base}}
    \State $curv \gets curv +  \tr(\nabla^2 f(z_i)) $
\EndFor
\State $curv\_estimate \gets curv / n_{iter}$\\
\Return $curv\_estimate$
\end{algorithmic}
\end{algorithm}

\subsection{Curvature based MIA}
\label{appndx:mia_curvature}
In this section, we present the detailed step by step description and pseudo code for the curvature based MIA (refer to Algorithm \ref{alg:mia_curvature})
\begin{enumerate}
    \item Initialization (Lines 3-4):
 Two empty sets, $curv_{in}$ and $curv_{out}$, are initialized. These sets will store the curvature scores for the shadow models trained with and without the target example $z_i$.
    
    \item Shadow Model Training (Lines 5-12):
    For a specified number of iterations $N$, the algorithm performs the following steps:
        \begin{itemize}
     
                    \item Sampling Shadow Dataset (Line 6): A shadow dataset $\mathcal{D}_{attack}$ is sampled from the data distribution $\mathcal{D}$.
                    \item Training IN Shadow Model (Line 7): A shadow model $g_{in}^{n}$ is trained on the shadow dataset augmented with the target example, $\mathcal{D}_{attack} \cup \set{z_i}$. The curvature score for this model, denoted as $curv_{in}^{n}$, is computed using Algorithm \ref{alg:pseudo-zo-curv}.
                    \item Collecting IN Curvature Scores (Line 8): The curvature score for the training set $\mathcal{D}_{attack} \cup \set{z_i}$ of this model $curv_{in}^{n}$ is added to the set $curv_{in}$.
                    \item Training OUT Shadow Model (Line 9): Another shadow model $g_{out}^{n}$ is trained on the shadow dataset excluding the target example, $\mathcal{D}_{attack} \setminus \set{z_i}$. The curvature score for this model, denoted as $curv_{out}^{n}$, is also computed using Algorithm \ref{alg:pseudo-zo-curv}.
                    \item Collecting OUT Curvature Scores (Line 10): The curvature score for the training set $\mathcal{D}_{attack} \setminus \set{z_i}$ of this model $curv_{out}^{n}$ is added to the set $curv_{out}$.
                \end{itemize}

    \item Model Parameter Estimation (Lines 13-16):    After training the shadow models, the algorithm calculates the mean ($\mu_{in}$, $\mu_{out}$) and variance ($\sigma_{in}^2$, $\sigma_{out}^2$) of the collected curvature scores for both the `IN' and `OUT' models.

    \item Target Model Query (Line 17): The curvature score for the target model $g_{target}$ with respect to the target example $z_i$, denoted as $curv_{target}$, is computed using Algorithm \ref{alg:pseudo-zo-curv}.

    \item Likelihood Ratio Test (Line 18-19): The likelihood ratio $P_{in}$ is calculated. This ratio compares the probability of the observed curvature score under the `IN' distribution against the `OUT'  distribution and is returned.
    For the `NLL' version $P_{in}$ us given by:
    \begin{align}
        P_{in}^{NLL} =  \log\left[P\left(curv_{target}  ~\lvert~ \mathcal{N}(\mu_{out}, \sigma_{out}^2)\right)\right] - \log\left[P\left(curv_{target} ~\lvert~ \mathcal{N}(\mu_{in} \sigma_{in}^2)\right)\right]
    \end{align}
\end{enumerate}
\begin{algorithm}
\caption{Membership Inference Attack using Input Loss Curvature Scores}
\label{alg:mia_curvature}
\begin{algorithmic}[1]
\Require Target model $g_{target}$, target example $z_i = (x_i, y_i)$, data distribution $\mathcal{D}$

\State $curv_{in} \gets \set{}$
\State $curv_{out} \gets \set{}$

\For{$n$ in $N$}
    \State $\mathcal{D}_{attack} \leftarrow \mathcal{D}$ \hfill\Comment{Sample a shadow dataset}
    \State $g_{in}^{n} \leftarrow \text{Train on  }\mathcal{D}_{attack} \cup \set{z_i}$. \hfill\Comment{Train in shadow model}
    \State $curv_{in}^{n} \gets $ Use Algorithm \ref{alg:pseudo-zo-curv}
    \State $curv_{in} \leftarrow curv_{in} \cup curv_{in}^{n}$
    \State $g_{out}^{n} \leftarrow \text{Train on  }\mathcal{D}_{attack} \setminus \set{z_i}$. \hfill\Comment{Train out shadow model}
    \State $curv_{out}^{n} \gets $ Use Algorithm \ref{alg:pseudo-zo-curv}
    \State $curv_{out} \leftarrow curv_{out} \cup curv_{out}^{n}$
\EndFor

\State $\mu_{in} \leftarrow \text{mean}(curv_{in})$
\State $\mu_{out} \leftarrow \text{mean}(curv_{out})$
\State $\sigma_{in}^2 \leftarrow \text{var}(curv_{in})$
\State $\sigma_{out}^2 \leftarrow \text{var}(curv_{out})$

\State $curv_{target} \gets $ Use Algorithm \ref{alg:pseudo-zo-curv} with $g_{target}$ and $z_i$. 
\State $P_{in} = \cfrac{P\left(curv_{target} ~\lvert~ \mathcal{N}(\mu_{in}, \sigma_{in}^2)\right)}{P\left(curv_{target}  ~\lvert~ \mathcal{N}(\mu_{out}, \sigma_{out}^2)\right)}$\\
\Return $P_{in}$

\end{algorithmic}
\end{algorithm}

\subsection{Proof of Theorem \ref{th:p_dkl}}
\label{appndx:th_p_dkl}
Consider an $\epsilon-$DP algorithm and $S, S^{\setminus i}$ such that $||S-S^{\setminus i} || = 1$. Next let $\mathcal{R} \subset \mathrm{Range}(\mathcal{A})$ such that $\mathcal{R} = \{h ~\lvert~ h(x_i) = y_i\}$. Since $\mathcal{A}$ is $\epsilon$-differentially private, then it follows from the definition of differential privacy in Equation \ref{eq:dp_def} that
\begin{align}
\Pr_\phi[h^\phi_S \in \mathcal{R}] & \leq e^{\epsilon} \Pr_\phi[h^\phi_{S\setminus i} \in \mathcal{R}] \\
\implies p(\phi, S, z_i) &\leq e^\epsilon p(\phi, S^{\setminus i}, z_i)
\label{eq:dp_priv_output}
\end{align}
Since $p(\phi, S, z_i) = \Pr_\phi[h^\phi_S \in \mathcal{R}]$ and denotes the probability density function of the neural network's output probability. Now, we use the definition of KL divergence
\begin{align*}
    \KL\left( p(\phi, S, z_i)~||~p(\phi, S^{\setminus i}, z_i) \right) &= \Ex_{\phi}\left[ \log \left( \cfrac{p(\phi, S, z_i)}{p(\phi, S^{\setminus i}, z_i)} \right)\right]\\ 
     &\leq \Ex_{\phi}\left[ \log e^{\epsilon} \right] \quad\text{From Equation \ref{eq:dp_priv_output}}\\
     \KL\left( p(\phi, S, z_i)~||~p(\phi, S^{\setminus i}, z_i) \right) &\leq \epsilon \quad \blacksquare
\end{align*}

\subsection{Lemma \ref{lm:curv_upper_bound}}

\begin{lemma}[Upper bound on Input Loss Curvature] \label{th:pr_curv}  Let  $\mathcal{A}$ be a randomized algorithm which is $\epsilon$-differentially private and the assumptions of error stability \ref{as:stability}, generalization \ref{as:gen}, and uniform model bias \ref{as:model_bias} hold. Further, assume $0 \leq \ell \leq L$. Then for two adjacent datasets $S, S^{\setminus i} \sim \mathcal{D}$ with a probability at least $1-\delta$ we have 
% \begin{align}
% \Ex_{z,\phi}[\Curv_{\phi}(z, S)] \leq& ~L(m+1)(1- e^{-\epsilon}) + (4m-1)\gamma \nonumber \\& ~ + 2(m-1)\Delta + \cfrac{\rho}{6} \Ex[\lVert \alpha \rVert^3].
% \label{eq:pr_curv}
% \end{align}
% \end{theorem}
\label{lm:curv_upper_bound}
\begin{align}
\Ex_{\phi}[\Curv_{\phi}(z_i, S^{\setminus i})] \leq& ~L(m(1- e^{-\epsilon}) + 1) + c_1 \label{eq:pr_curv}
% \\& ~ + 2(m-1)\Delta + \cfrac{\rho}{6} \Ex[\lVert \alpha \rVert^3]. 
\\
c_1 = (4m-1)\gamma &+ 2(m-1)\Delta + \cfrac{\rho}{6} \Ex[\lVert \alpha \rVert^3]
\end{align}
\end{lemma}

\textbf{Proof of Lemma \ref{lm:curv_upper_bound}}
For this proof, we use results from \citet{nesterov2006cubic}, we restate it here for convenience
\begin{lemma} If Lipschitz assumption \ref{as:hess_lip} on the Hessian of $\ell$ holds from \citet{nesterov2006cubic} we have
\begin{align}
|\ell(h, z_1) - \ell(h, z_2) - \langle \nabla \ell(h, z_2), z_1 - z_2\rangle  - \langle \nabla^2 \ell(h, z_2)(z_1 - z_2), z_1 - z_2\rangle | \leq \cfrac{\rho}{6} |z_1 - z_2|^3 
\end{align}
\label{lm:hess_lip}
\end{lemma}
From Lemma \ref{lm:hess_lip} we have
\begin{align*}
-\cfrac{\rho}{6} |z_1 - z_2|^3  \leq \ell(h, z_1) - \ell(h, z_2) - \langle \nabla \ell(h, z_2), z_1 - z_2\rangle  - \langle \nabla^2 \ell(h, z_2)(z_1 - z_2), z_1 - z_2\rangle \\ \ell(h, z_1) - \ell(h, z_2) - \langle \nabla \ell(h, z_2), z_1 - z_2\rangle  - \langle \nabla^2 \ell(h, z_2)(z_1 - z_2), z_1 - z_2\rangle \leq \cfrac{\rho}{6} |z_1 - z_2|^3
\end{align*}
This gives us a lower bound on $\ell(h, z_1)$   
\begin{align}
-\cfrac{\rho}{6} |z_1 - z_2|^3 + \ell(h, z_2) + \langle \nabla \ell(h, z_2), z_1 - z_2\rangle  + \langle \nabla^2 \ell(h, z_2)(z_1 - z_2), z_1 - z_2\rangle &\leq \ell(h, z_1)   \label{eq:loss_hess_lower}
\end{align}
Consider $z_j \in S$ such that $z_i = z_j + \alpha$ for some $j\neq i$ where $\alpha \in B_p(\upsilon)$ such that $\Ex[\alpha] = 0$ and $\Ex[\alpha^T \alpha] = 1$. Using the lower bound in Lemma A.2 from \citet{ravikumar2024unveiling} with $z_1 = z_i, z_2 = z_j$ we get
\begin{align}
\Ex_\phi[\ell(h_{S^{\setminus i}}^\phi, z_j)] -  \Ex_\phi[\ell(h^\phi_S, z_i)] &\leq m\beta + (4m-1)\gamma + 2(m-1)\Delta 
 \label{eq:pre_polyak}
\end{align}
Using the result from Lemma \ref{lm:hess_lip}
\begin{align}
- \cfrac{\rho}{6} \lVert\alpha\rVert^3 + \Ex_\phi[\ell(h_{S^{\setminus i}}^\phi, z_i)] + \Ex_\phi[\langle \nabla \ell(h_{S^{\setminus i}}^\phi, z_i), \alpha\rangle] & \nonumber \\+ \Ex_\phi[\langle \nabla^2 \ell(h_{S^{\setminus i}}^\phi, z_i)\alpha, \alpha\rangle] - \Ex_\phi[\ell(h^\phi_S, z_i)] & ~\leq m\beta + (4m-1)\gamma + 2(m-1)\Delta
\label{eq:post_lm_polyak}
\end{align}
% \begin{align*}
% \Ex_\phi[\ell(h_S^\phi, z_j)] + \Ex_\phi[\langle \nabla \ell(h_S^\phi, z_j), \alpha\rangle] + \Ex_\phi[\langle \nabla^2 \ell(h_S^\phi, z_i)\alpha, \alpha\rangle]  - \Ex_\phi[\ell(h^\phi_{S^{\setminus i}}, z_j)]  \leq m\beta &+ (4m-1)\gamma + 2(m-1)\Delta + \cfrac{\rho}{6} \lVert \alpha \rVert^3 
% \end{align*}

Taking Expectation over $\alpha$ and since the mean of $\alpha = 0$ we have
\begin{align*}
\Ex_\phi[\ell(h_{S^{\setminus i}}^\phi, z_i)] + \Ex_{\alpha, \phi}[\langle \nabla \ell(h_{S^{\setminus i}}^\phi, z_i), \alpha\rangle] &+ \Ex_{\alpha,\phi}[\langle \nabla^2 \ell(h_{S^{\setminus i}}^\phi, z_i)\alpha, \alpha\rangle] \\ - \Ex_\phi[\ell(h^\phi_{S}, z_i)]  &\leq m\beta + (4m-1)\gamma + 2(m-1)\Delta + \cfrac{\rho}{6} \lVert \alpha \rVert^3
\end{align*}
Note that we can change the order of expectation using Fubini's theorem
\begin{align*}
\Ex_\phi[\ell(h_{S^{\setminus i}}^\phi, z_i)] + \Ex_{\phi, \alpha}[\langle \nabla^2 \ell(h_{S^{\setminus i}}^\phi, z_i)\alpha, \alpha\rangle]  - \Ex_\phi[\ell(h^\phi_S, z_i)]  &\leq m\beta + (4m-1)\gamma + 2(m-1)\Delta + \cfrac{\rho}{6} \lVert \alpha \rVert^3 \\
\Ex_\phi[\ell(h_{S^{\setminus i}}^\phi, z_i)] + \Ex_{\phi}[\mathrm{tr}(\nabla^2 \ell(h^\phi_{S^{\setminus i}}, z_i))]  - \Ex_\phi[\ell(h^\phi_S, z_i)] &\leq m\beta + (4m-1)\gamma + 2(m-1)\Delta + \cfrac{\rho}{6} \lVert \alpha \rVert^3 \\
\Ex_{\phi}[\mathrm{tr}(\nabla^2 \ell(h^\phi_{S^{\setminus i}}, z_i))] \leq m\beta + (4m-1)\gamma + 2(m-1)\Delta ~+&~ \cfrac{\rho}{6} \Ex[\lVert \alpha \rVert^3] + \Ex_\phi[\ell(h_S^\phi, z_i)]  -  \Ex_\phi[\ell(h^\phi_{S^{\setminus i}}, z_j)]\\
\Ex_{\phi}[\mathrm{tr}(\nabla^2 \ell(h^\phi_S, z_i))] \leq m\beta + (4m-1)\gamma + 2(m-1)\Delta ~+&~ \cfrac{\rho}{6} \Ex[\lVert \alpha \rVert^3] + L
\end{align*}
From \citet{ravikumar2024unveiling} Lemma 5.2 we have $\beta \leq L(1 - e^{-\epsilon})$. Thus we have the upper bound:
\begin{align}
\Ex_{\phi}[\Curv_{\phi}(z, S^{\setminus i})] \leq& ~L(m(1- e^{-\epsilon}) + 1) + c_1 
% \\& ~ + 2(m-1)\Delta + \cfrac{\rho}{6} \Ex[\lVert \alpha \rVert^3]. 
\\
c_1 = (4m-1)\gamma &+ 2(m-1)\Delta + \cfrac{\rho}{6} \Ex[\lVert \alpha \rVert^3] \quad \blacksquare
\end{align}

\subsection{Proof of Theorem \ref{th:pc_dkl}}
\label{appndx:th_pc_dkl}
This proof uses results from Lemma \ref{lm:curv_upper_bound} provided above. For the proof of Theorem \ref{th:pc_dkl} let $p_c(\phi, S, z_i)$ be the curvature probability mass function, then using the Gaussian model discussed in Section \ref{sec:theory} we model it as a Gaussian. Thus, we can write 
\begin{align}
    p_c(\phi, S, z_i) &= \cfrac{1}{\sigma_{S}\sqrt{2 \pi}} ~ e^{-\cfrac{(\phi - \mu_S)^2}{2 \sigma_{S}^2}}\\
    p_c(\phi, S^{\setminus i}, z_i) &= \cfrac{1}{\sigma_{S^{\setminus i}}\sqrt{2 \pi}} ~ e^{-\cfrac{(\phi - \mu_{S^{\setminus i}})^2}{2 \sigma_{S^{\setminus i}}^2}}
\end{align}
Now consider the $\KL(p_c(\phi, S, z_i)~||~p_c(\phi, S^{\setminus i}, z_i))$
\begin{align}
    \KL(p_c(\phi, S, z_i)~||~p_c(\phi, S^{\setminus i}, z_i)) &= \Ex_{\phi}\left[ \log \left( \cfrac{p_c(\phi, S, z_i)}{p_c(\phi, S^{\setminus i}, z_i))} \right) \right]\\
    &= \Ex_{\phi}\left[ -\cfrac{(\phi - \mu_S)^2}{2 \sigma^2} + \cfrac{(\phi - \mu_{S^{\setminus i}})^2}{2 \sigma^2}\right] \quad \text{Assume}~ \sigma = \sigma_{S^{\setminus i}} = \sigma_{S} \\
    &= \Ex_{\phi}\left[\cfrac{-\phi^2 - \mu_S^2 + 2\phi \mu_S + \phi^2 + \mu_{S^{\setminus i}}^2 - 2\phi \mu_{S^{\setminus i}}}{2 \sigma^2} \right] \\
    &= \Ex_{\phi}\left[\cfrac{\mu_{S^{\setminus i}}^2 - \mu_S^2}{2 \sigma^2} \right] \\
    &= \cfrac{\mu_{S^{\setminus i}}^2 - \mu_S^2}{2 \sigma^2} \label{eq:model_diff}
\end{align}
To upper bound Equation \ref{eq:model_diff}, we need an upper bound on $\mu_{S^{\setminus i}}$ and a lower bound on $\mu_S$. The lower bound on $\mu_S^2 = 0$

For ease of notation let's define $\beta_{max} := L(m(1- e^{-\epsilon}) + 1)$. Notice that the upper bound of per sample $\mu_{S^{\setminus i}} = \Ex_{\phi}[\Curv_{\phi}(z, S^{\setminus i})]$.
Using the result from Lemma \ref{lm:curv_upper_bound} in Equation \ref{eq:model_diff} we have
% \begin{align}
%     &\leq \cfrac{(\beta_{max}+ c_2)^2 - \left(\KL\left( p(\pmb{\phi}, S, z_i)~||~p(\pmb{\phi}, S^{\setminus i}, z_i) \right) - c_1 \right)^2}{2 \sigma^2}\\
%     &\leq(\beta_{max}+ c_1)^2 - \left(\KL\left( p(\pmb{\phi}, S, z_i)~||~p(\pmb{\phi}, S^{\setminus i}, z_i \right) - c_1 \right)^2\\
%     &\leq \beta_{max}^2 + c_1^2 + 2\beta_{max} c_1 - \epsilon^2 - c_1^2 + 2 \KL c_1\\
%     &\leq \beta_{max}^2  - \epsilon^2\\
%     &\leq L (m+1)(1-e^{-\epsilon}) - \epsilon^2
% \end{align}
% For $\KL(p_c(\pmb{\phi}, S, z_i)~||~p_c(\pmb{\phi}, S^{\setminus i}, z_i)) > \KL\left( p(\pmb{\phi}, S, z_i)~||~p(\pmb{\phi}, S^{\setminus i}, z_i) \right)$
% We have
% \begin{align*}
%     L (m+1)(1-e^{-\epsilon}) - \epsilon^2 &\geq \epsilon\\
%     m &\geq \cfrac{\epsilon + \epsilon^2 }{L(1-e^{-\epsilon})} - 1
% \end{align*}
\begin{align}
    \KL(p_c(\phi, S, z_i)~||~p_c(\phi, S^{\setminus i}, z_i)) &\leq \cfrac{(\beta_{max}+ c_1)^2 - 0^2}{2 \sigma^2}\\
    &\leq \cfrac{(\beta_{max} + c_1)^2}{2 \sigma^2}\\
    &\leq \cfrac{\left[L(m(1- e^{-\epsilon}) + 1) + c_1\right]^2}{2 \sigma^2}\quad \blacksquare
    % &\leq(\beta_{max}+ c_1)^2 - \left(\KL\left( p(\pmb{\phi}, S, z_i)~||~p(\pmb{\phi}, S^{\setminus i}, z_i \right) - c_1 \right)^2\\
    % &\leq \beta_{max}^2 + c_1^2 + 2\beta_{max} c_1 - \epsilon^2 - c_1^2 + 2 \KL c_1\\
    % &\leq \beta_{max}^2  - \epsilon^2\\
    % &\leq L (m+1)(1-e^{-\epsilon}) - \epsilon^2
\end{align}

\subsection{Proof of Theorem \ref{th:m_limit}}
\label{appndx:th_m_limit}
Assuming the bounds of Theorem \ref{th:p_dkl} and Theorem \ref{th:pc_dkl} are tight, then using curvature score will outperform neural network probability scores if:
\begin{align}
    \KL(p_c(\phi, S, z_i)~||~p_c(\phi, S^{\setminus i}, z_i)) &> \KL(p(\phi, S, z_i)~||~p(\phi, S^{\setminus i}, z_i)) \label{eq:perf_req}
\end{align}
\begin{align*}
    \cfrac{\left[L(m(1- e^{-\epsilon}) + 1) + c_1\right]^2}{2 \sigma^2} &> \epsilon\\
    \cfrac{\left[Lm(1- e^{-\epsilon}) + c\right]^2}{2 \sigma^2} &> \epsilon \quad; c = c_1 + L\\
    L^2(1-e^{-\epsilon})^2 m^2 + 2Lc(1-e^{-\epsilon})m + c^2 - 2\sigma^2\epsilon &> 0
\end{align*}
Using the quadratic formula we have the roots given by:
\begin{align*}
    &= \cfrac{-2Lc(1-e^{-\epsilon}) \pm \sqrt{4L^2c^2(1-e^{-\epsilon})^2 - 4L^2(1-e^{-\epsilon})^2(c^2 - 2\sigma^2\epsilon) }  }{2L^2(1-e^{-\epsilon})^2}\\
    &=\cfrac{-2Lc \pm \sqrt{4L^2c^2 - 4L^2(c^2 - 2\sigma^2\epsilon) }  }{2L^2(1-e^{-\epsilon})}\\
    &=\cfrac{-Lc \pm \sqrt{L^2c^2 - L^2(c^2 - 2\sigma^2\epsilon) }  }{L^2(1-e^{-\epsilon})}\\
    &=\cfrac{-c \pm \sqrt{c^2 - (c^2 - 2\sigma^2\epsilon) }  }{L(1-e^{-\epsilon})}\\
    &=\cfrac{-c \pm \sqrt{2\sigma^2\epsilon}  }{L(1-e^{-\epsilon})}\\
\end{align*}
Assuming both roots are real, the larger root is $\cfrac{\sqrt{2\sigma^2\epsilon} - c }{L(1-e^{-\epsilon})}$
Thus when $m > \cfrac{\sqrt{2\sigma^2\epsilon} - c }{L(1-e^{-\epsilon})}$  Equation \ref{eq:perf_req} is satisfied. $\blacksquare$

\subsection{Input Augmentations Ablation}
\label{appndx:aug_results}
In this section we present the results of using 1 augmentation (only the image) vs 2 augmentations (image + mirror) in Tables \ref{tab:imagenet_results_aug}, \ref{tab:cifar100_results_aug} and \ref{tab:cifar10_results_aug} for ImageNet, CIFAR100 and CIFAR10 respectively. For these results, we used 64 shadow models for CIFAR10 and CIFAR100 and 52 shadow models for ImageNet.

% ImageNet Table
\begin{table}[hbt!]
\renewcommand*{\arraystretch}{1.0}
\centering
\begin{adjustbox}{width=0.75\textwidth}
\begin{tabular}{lcccc}
\toprule
\multirow{3}{*}{Method} & \multicolumn{4}{c}{ImageNet} \\
\cmidrule{2-5}
 & \multicolumn{2}{c}{1 Augmentation} & \multicolumn{2}{c}{2 Augmentations} \\
\cmidrule{2-3} \cmidrule{4-5}
 & Bal. Acc. & AUROC & Bal. Acc. & AUROC \\
\midrule
Curv ZO NLL (Ours) & 68.09 {\scriptsize$\pm$ 0.07} & 75.84 {\scriptsize$\pm$ 0.08} & 69.16 {\scriptsize$\pm$ 0.08} & 77.45 {\scriptsize$\pm$ 0.09} \\
Curv ZO LR (Ours) & 67.59 {\scriptsize$\pm$ 0.05} & 70.62 {\scriptsize$\pm$ 0.04} & 68.76 {\scriptsize$\pm$ 0.04} & 72.28 {\scriptsize$\pm$ 0.04} \\
% Curv ZO 1-NLL (Ours) & 64.20 {\scriptsize$\pm$ 0.12} & 71.94 {\scriptsize$\pm$ 0.12} & - & - \\
\cite{carlini2022membership} & 65.11 {\scriptsize$\pm$ 0.03} & 71.96 {\scriptsize$\pm$ 0.04} & 66.14 {\scriptsize$\pm$ 0.01} & 73.46 {\scriptsize$\pm$ 0.02} \\
\bottomrule
\end{tabular}
\end{adjustbox}
\vspace{2mm}
\caption{Comparison of the proposed curvature score-based MIA with prior methods on ImageNet dataset with 1 and 2 augmentations. Results reported are the mean $\pm$ std obtained over 3 seeds.}
\label{tab:imagenet_results_aug}
\end{table}

% CIFAR100 Table
\begin{table}[hbt!]
\renewcommand*{\arraystretch}{1.0}
\centering
\begin{adjustbox}{width=0.75\textwidth}
\begin{tabular}{lcccc}
\toprule
\multirow{3}{*}{Method} & \multicolumn{4}{c}{CIFAR100} \\
\cmidrule{2-5}
 & \multicolumn{2}{c}{1 Augmentation} & \multicolumn{2}{c}{2 Augmentations} \\
\cmidrule{2-3} \cmidrule{4-5}
 & Bal. Acc. & AUROC & Bal. Acc. & AUROC \\
\midrule
Curv ZO NLL (Ours) & 83.97 {\scriptsize$\pm$ 0.23} & 92.72 {\scriptsize$\pm$ 0.23} & 84.47 {\scriptsize$\pm$ 0.21} & 93.49 {\scriptsize$\pm$ 0.18} \\
Curv ZO LR (Ours) & 77.88 {\scriptsize$\pm$ 0.21} & 88.10 {\scriptsize$\pm$ 0.12} & 80.48 {\scriptsize$\pm$ 0.10} & 90.15 {\scriptsize$\pm$ 0.04} \\
\cite{carlini2022membership} & 80.73 {\scriptsize$\pm$ 0.16} & 87.69 {\scriptsize$\pm$ 0.18} & 81.55 {\scriptsize$\pm$ 0.13} & 88.89 {\scriptsize$\pm$ 0.16} \\
\bottomrule
\end{tabular}
\end{adjustbox}
\vspace{2mm}
\caption{Comparison of the proposed curvature score-based MIA with prior methods on CIFAR100 dataset with 1 and 2 augmentations. Results reported are the mean $\pm$ std obtained over 3 seeds.}
\label{tab:cifar100_results_aug}
\end{table}

\begin{table}[hbt!]
\renewcommand*{\arraystretch}{1.0}
\centering
\begin{adjustbox}{width=0.75\textwidth}
\begin{tabular}{lcccccccccccc}
\toprule
\multirow{3}{*}{Method} & \multicolumn{4}{c}{CIFAR10} \\
\cmidrule{2-5}
 & \multicolumn{2}{c}{1 Augmentation} & \multicolumn{2}{c}{2 Augmentations} \\
\cmidrule{2-3} \cmidrule{4-5}
 & Bal. Acc. & AUROC & Bal. Acc. & AUROC \\
\midrule
Curv ZO NLL (Ours) & 61.03 {\scriptsize$\pm$ 0.84} & 67.56 {\scriptsize$\pm$ 1.23} & 61.92 {\scriptsize$\pm$ 0.87} & 68.82 {\scriptsize$\pm$ 1.30} \\
Curv ZO LR (Ours) & 54.54 {\scriptsize$\pm$ 0.19} & 58.06 {\scriptsize$\pm$ 0.41} & 55.00 {\scriptsize$\pm$ 0.17} & 58.89 {\scriptsize$\pm$ 0.38} \\
\cite{carlini2022membership} & 56.96 {\scriptsize$\pm$ 0.27} & 60.05 {\scriptsize$\pm$ 0.34} & 58.23 {\scriptsize$\pm$ 0.29} & 61.73 {\scriptsize$\pm$ 0.32} \\
\bottomrule
\end{tabular}
\end{adjustbox}
\vspace{2mm}
\caption{Comparison of the proposed curvature score-based MIA with prior methods on CIFAR10 dataset with 1 and 2 augmentations. Results reported are the mean $\pm$ std obtained over 3 seeds.}
\label{tab:cifar10_results_aug}
\end{table}

\subsection{Low FPR Results}
\label{appndx:tpr_fpr}
In this section, we present the performance of curvature based MIA compared with prior works using TPR (true positive rate) at very low FPR percentages. The results are presented when using 1 augmentation (only the image) and 2 augmentations (image + mirror) in Table \ref{tab:cifar100_tpr_fpr}. For these results, we use 64 shadow models for all the results in Table \ref{tab:cifar100_tpr_fpr} (for methods that use shadow models). The results for \citet{carlini2022membership} are a little lower than the one reported by \citet{carlini2022membership} in TABLE I of their paper. However, note that the authors of \citet{carlini2022membership} report the results with 256 shadow models while we report it using 64 shadow models. This highlights the benefits of curvature based approach, it achieves similar performance to \citet{carlini2022membership} with 256 shadow models but with only 64 shadow models thus needs significantly less shadow models to achieve similar performance.

\textbf{Takeaways:} The parametric curvature LR method has the best TPR at very low FPR and makes efficient use of shadow models. 

\begin{table}[hbt!]
\renewcommand*{\arraystretch}{1.2}
\begin{adjustbox}{width=\textwidth}
\centering
\begin{tabular}{lcccc}
\toprule
\multirow{2}{*}{Method} & \multicolumn{2}{c}{1 Augmentation} & \multicolumn{2}{c}{2 Augmentations} \\
\cmidrule{2-3} \cmidrule{4-5}
 & TPR @ 0.1\% FPR & TPR @ 0.01\% FPR & TPR @ 0.1\% FPR & TPR @ 0.01\% FPR \\
\midrule
\textbf{Curv ZO LR (Ours)} & \textbf{21.07 {\scriptsize$\pm$ 0.80}} & \textbf{15.74 {\scriptsize$\pm$ 3.12}} & \textbf{23.92 {\scriptsize$\pm$ 0.92}} & \textbf{17.17 {\scriptsize$\pm$ 1.87}} \\
Curv ZO NLL (Ours) & 6.42 {\scriptsize$\pm$ 0.62} & 0.05 {\scriptsize$\pm$ 0.04} & 8.29 {\scriptsize$\pm$ 1.52} & 0.10 {\scriptsize$\pm$ 0.14} \\
\cite{carlini2022membership} & 15.52 {\scriptsize$\pm$ 0.54} & 5.02 {\scriptsize$\pm$ 1.70} & 15.80 {\scriptsize$\pm$ 0.49} & 6.56 {\scriptsize$\pm$ 1.13} \\
\cite{sablayrolles2019white} & 10.50 {\scriptsize$\pm$ 0.59} & 4.30 {\scriptsize$\pm$ 0.53} & 10.50 {\scriptsize$\pm$ 0.59} & 4.30 {\scriptsize$\pm$ 0.53} \\
\cite{watson2021importance} & 6.62 {\scriptsize$\pm$ 0.23} & 2.95 {\scriptsize$\pm$ 0.18} & 6.62 {\scriptsize$\pm$ 0.23} & 2.95 {\scriptsize$\pm$ 0.18} \\
\cite{ye2022enhanced} & 6.63 {\scriptsize$\pm$ 0.52} & 2.86 {\scriptsize$\pm$ 0.42} & 6.63 {\scriptsize$\pm$ 0.52} & 2.86 {\scriptsize$\pm$ 0.42} \\
\cite{song2021systematic} & 1.21 {\scriptsize$\pm$ 0.37} & 1.21 {\scriptsize$\pm$ 0.37} & 1.21 {\scriptsize$\pm$ 0.37} & 1.21 {\scriptsize$\pm$ 0.37} \\
\cite{yeom2018privacy} & 0.07 {\scriptsize$\pm$ 0.01} & 0.01 {\scriptsize$\pm$ 0.00} & 0.07 {\scriptsize$\pm$ 0.01} & 0.01 {\scriptsize$\pm$ 0.00} \\
\bottomrule
\end{tabular}
\end{adjustbox}
\vspace{2mm}
\caption{Comparison of TPR @ 0.1\% FPR and TPR @ 0.01\% FPR for CIFAR100 dataset with 1 and 2 augmentations. Results reported are the mean $\pm$ std obtained over 3 seeds and using 64 shadow models.}
\label{tab:cifar100_tpr_fpr}
\end{table}

\subsection{Broader Impact}
\label{appndx:borader_impact}
The development and analysis of input loss curvature in deep neural networks presented in this work have significant implications for both the academic and practical fields of machine learning and privacy. By introducing a novel black-box membership inference attack that leverages input loss curvature, this research advances the state-of-the-art in privacy testing for machine learning models. 
% This new technique enhances our understanding of how neural networks memorize training data and offers a more effective method for detecting if a particular data point was used during training.

From a societal perspective, the ability to detect and mitigate membership inference attacks is crucial for maintaining user privacy in machine learning applications. This work helps pave the way for more robust privacy-preserving techniques, ensuring that sensitive information is protected against unauthorized inference. Furthermore, the insights gained from the relationship between input loss curvature and memorization can guide the development of more secure machine learning models, which is particularly important as these models are increasingly deployed in sensitive domains such as healthcare, finance, and personal data processing.

Additionally, this research highlights the potential of using subsets of training data as a defense mechanism against membership inference attacks. By identifying that training on certain sized subsets can improve resistance to these attacks, this work offers practical guidance for model training practices that can enhance privacy without significantly compromising performance.

\subsection{Limitations}
\label{appndx:limitations}
In this paper we presented the a theoretical analysis and experimental evidence for improved MIA using input loss curvature. The theoretical and empirical evidence also shows that certain sized subsets of the training set may provide defense against membership inference attacks. However, as mentioned in the paper, the MIA performance improvement occurs only above a certain training dataset size below which the non-parametric model of `Curv NLL' works better and is not explained by the theoretical analysis. Similar to techniques that use shadow models such as by \citet{shokri2017membership}, \citet{carlini2022membership} we need to train shadow models, which can be computationally expensive. Further, the method described requires more queries at minimum $4\times$ more than \citet{carlini2022membership}. We believe these limitations can be addressed by follow up research. 

\subsection{Reproducibility Details}
\label{appndx:reprod_docu}
In this section we present additional details for reproducing our results.

\textbf{Training.}
For experiments that use private models, we use the Opacus library \citep{opacus} to train ResNet18 models for 20 epochs till the privacy budget is reached. We use DP-SGD \citep{abadi2016deep} with the maximum gradient norm set to $1.0$ and privacy parameter $\delta=1\times 10^{-5}$.
% Training used the RMSProp \cite{hinton2012neural} optimizer with
The initial learning rate was set to $0.001$. The learning rate is decreased by $10$ at epochs $12$ and $16$ with a batch size of $128$. 

For shadow model training on CIFAR10 and CIFAR100 we trained on 50\% randomly sampled subset of the data for 300 epochs with a batch size of 512 for CIFAR100 and 256 for CIFAR10. We used SGD optimizer with the initial learning rate set to 0.1, weight decay of $1 \times 10^{-4}$ and momenutm of 0.9. The learning rate was decayed by 0.1 at $180^{th}$ and $240^{th}$ epoch. For ImageNet we used pre-trained models from \citet{feldman2020neural} as shadow models which were trained on a 70\% subset of ImageNet. For both CIFAR10 and CIFAR100 datasets, we used the following sequence of data augmentations for training: resize ($32 \times 32$), random crop, and random horizontal flip, this is followed by normalization.

\textbf{Testing.} 
During testing we used resize followed by normalization. We used two augmentations the original image and its mirror. The number of augmentations used are specified in the corresponding experiment section. When using pre-trained models from \citet{feldman2020neural} we validated the accuracy of the models before performing experiments. 

\textbf{Compute Resources.} All of the experiments were performed on a heterogeneous compute cluster consisting of 9 1080Ti's, 6 2080Ti's and 4 A40 NVIDIA GPUs, with a total of 100 CPU cores and a combined 1.2 TB of main system memory. However, the results can be replicated with a single GPU with 11GB of VRAM. 

\textbf{Hyperparameters.} Our code uses 2 hyper parameters for zero-order input loss curvature estimation. The $n_{iter}$ and $h$ in Algorithm \ref{alg:pseudo-zo-curv}. We used $n_{iter} = 10$ and $h = 0.001$. To improve reproducibility, we have provided the code in the supplementary material.

\subsection{Licenses for Assets Used}
\label{appndx:asset_license}
For each of the assets used we present the licenses below we also have provided the correct citation in the main paper as well as here for convenience.

\begin{enumerate}
    \item ImageNet \citep{ILSVRC15}: Terms of access available at \url{https://image-net.org/download.php}
    \item CIFAR10 \citep{krizhevsky2009learning}: Unknown / No license provided
    \item CIFAR100 \citep{krizhevsky2009learning}: Unknown / No license provided
    \item Pre-trained ImageNet models and code: We used pre-trained ImageNet models from \citet{feldman2020neural} which is licensed under Apache 2.0 \url{https://github.com/google-research/heldout-influence-estimation/blob/master/LICENSE}.
    \item Baseline methods: We re-implemented the baseline methods hence is provided with along with our code which is distributed under the MIT License.
    \item Opacus \citep{opacus}: Licensed under Apache 2.0, \url{https://github.com/pytorch/opacus/blob/main/LICENSE}.
    \item Pytorch \citep{Ansel_PyTorch_2_Faster_2024}: Custom BSD-style license available at \url{https://github.com/pytorch/pytorch/blob/main/LICENSE}.
    \item ResNet Model Architecture \citep{he2016deep}: MIT license available at \url{https://github.com/kuangliu/pytorch-cifar/blob/master/LICENSE}
\end{enumerate}

\newpage
\section*{NeurIPS Paper Checklist}

\begin{enumerate}

\item {\bf Claims}
    \item[] Question: Do the main claims made in the abstract and introduction accurately reflect the paper's contributions and scope?
    \item[] Answer: \answerYes{} % \answerTODO{} % Replace by \answerYes{}, \answerNo{}, or \answerNA{}.
    \item[] Justification: The abstract clearly outlines the exploration of input loss curvature in deep neural networks and the development of a theoretical framework for train-test distinguishability. It introduces a novel black-box membership inference attack (MIA) utilizing input loss curvature and the use of subsets of training data as a mechanism to defend against shadow model based MIA. These claims are substantiated through theoretical analysis and experimental validation, as presented in the paper.

\item {\bf Limitations}
    \item[] Question: Does the paper discuss the limitations of the work performed by the authors?
    \item[] Answer: \answerYes{} % Replace by \answerYes{}, \answerNo{}, or \answerNA{}.
    \item[] Justification: We discuss the limitations of the work in Appendix \ref{appndx:limitations}, where we note the increased computation requirement imposed by zero-order curvature calculation and other limitations of the proposed methods.

\item {\bf Theory Assumptions and Proofs}
    \item[] Question: For each theoretical result, does the paper provide the full set of assumptions and a complete (and correct) proof?
    \item[] Answer: \answerYes{} % Replace by \answerYes{}, \answerNo{}, or \answerNA{}.
    \item[] Justification: All the assumptions are stated in the Theorems and discussed prior to their utilization in the main paper. The complete proof has also been provided in the Appendix of the paper, specifically Appendix \ref{appndx:th_p_dkl}, \ref{appndx:th_pc_dkl} and \ref{appndx:th_m_limit}.

    \item {\bf Experimental Result Reproducibility}
    \item[] Question: Does the paper fully disclose all the information needed to reproduce the main experimental results of the paper to the extent that it affects the main claims and/or conclusions of the paper (regardless of whether the code and data are provided or not)?
    \item[] Answer: \answerYes{} % Replace by \answerYes{}, \answerNo{}, or \answerNA{}.
    \item[] Justification: We include all the training details for the models in Appendix \ref{appndx:reprod_docu}. We also provide the pseudo-code for the proposed technique in Appendix \ref{appndx:mia_curvature} along with all the hyper parameter details (see Appendix \ref{appndx:reprod_docu}).  Further, we also provide the code in the supplementary material.

\item {\bf Open access to data and code}
    \item[] Question: Does the paper provide open access to the data and code, with sufficient instructions to faithfully reproduce the main experimental results, as described in supplemental material?
    \item[] Answer: \answerYes{} %\answerTODO{} % Replace by \answerYes{}, \answerNo{}, or \answerNA{}.
    \item[] Justification: The datasets used are publicly available and are the standard datasets used for vision tasks. Our code is included in the supplementary material and will be released publicly after the conference deadline. The code replicates the results of our experiments.

\item {\bf Experimental Setting/Details}
    \item[] Question: Does the paper specify all the training and test details (e.g., data splits, hyperparameters, how they were chosen, type of optimizer, etc.) necessary to understand the results?
    \item[] Answer: \answerYes{} % Replace by \answerYes{}, \answerNo{}, or \answerNA{}.
    \item[] Justification: The training details are reported in Section \ref{sec:exp_setup} of the paper and more details are provided in Appendix \ref{appndx:reprod_docu}.

\item {\bf Experiment Statistical Significance}
    \item[] Question: Does the paper report error bars suitably and correctly defined or other appropriate information about the statistical significance of the experiments?
    \item[] Answer:  \answerYes{} %\answerTODO{} % Replace by \answerYes{}, \answerNo{}, or \answerNA{}.
    \item[] Justification: All reported results are obtained by averaging over 3 seeds and reported with mean and standard deviation values. %\justificationTODO{}

\item {\bf Experiments Compute Resources}
    \item[] Question: For each experiment, does the paper provide sufficient information on the computer resources (type of compute workers, memory, time of execution) needed to reproduce the experiments?
    \item[] Answer: \answerYes{} %\answerTODO{} % Replace by \answerYes{}, \answerNo{}, or \answerNA{}.
    \item[] Justification: For the experiments, we used a heterogeneous cluster whose details are mentioned in Appendix \ref{appndx:reprod_docu}.
    
\item {\bf Code Of Ethics}
    \item[] Question: Does the research conducted in the paper conform, in every respect, with the NeurIPS Code of Ethics \url{https://neurips.cc/public/EthicsGuidelines}?
    \item[] Answer: \answerYes{} % Replace by \answerYes{}, \answerNo{}, or \answerNA{}.
    \item[] Justification: The paper conforms to the code of ethics, we discus potential impacts of this research paper in Appendix \ref{appndx:borader_impact}.  % should we mention the safety, security, discrimination, etc from the website?

\item {\bf Broader Impacts}
    \item[] Question: Does the paper discuss both potential positive societal impacts and negative societal impacts of the work performed?
    \item[] Answer: \answerYes{} % Replace by \answerYes{}, \answerNo{}, or \answerNA{}.
    \item[] Justification: We provide a detailed discussion about the broader impact of the research presented in the paper in Appendix \ref{appndx:borader_impact}.
    
\item {\bf Safeguards}
    \item[] Question: Does the paper describe safeguards that have been put in place for responsible release of data or models that have a high risk for misuse (e.g., pretrained language models, image generators, or scraped datasets)?
    \item[] Answer: \answerNA{} % Replace by \answerYes{}, \answerNo{}, or \answerNA{}.
    \item[] Justification: We do not release data or models thus this paper does not pose such risks. %\justificationTODO{}

\item {\bf Licenses for existing assets}
    \item[] Question: Are the creators or original owners of assets (e.g., code, data, models), used in the paper, properly credited and are the license and terms of use explicitly mentioned and properly respected?
    \item[] Answer: \answerYes{} % Replace by \answerYes{}, \answerNo{}, or \answerNA{}.
    \item[] Justification: We provide the detailed list of assets used and the terms under which they were licensed in Appendix \ref{appndx:asset_license}. We have respected the licensing terms of the original owners.

\item {\bf New Assets}
    \item[] Question: Are new assets introduced in the paper well documented and is the documentation provided alongside the assets?
    \item[] Answer: \answerYes{} % Replace by \answerYes{}, \answerNo{}, or \answerNA{}.
    \item[] Justification: We release the code for reproducing the results under the MIT license and details of its usage are documented in Appendix \ref{appndx:reprod_docu} and the supplementary material README file.

\item {\bf Crowdsourcing and Research with Human Subjects}
    \item[] Question: For crowdsourcing experiments and research with human subjects, does the paper include the full text of instructions given to participants and screenshots, if applicable, as well as details about compensation (if any)? 
    \item[] Answer: \answerNA{} %\answerTODO{} % Replace by \answerYes{}, \answerNo{}, or \answerNA{}.
    \item[] Justification: The paper does not involve crowdsourcing or research with human subjects. % \justificationTODO{}

\item {\bf Institutional Review Board (IRB) Approvals or Equivalent for Research with Human Subjects}
    \item[] Question: Does the paper describe potential risks incurred by study participants, whether such risks were disclosed to the subjects, and whether Institutional Review Board (IRB) approvals (or an equivalent approval/review based on the requirements of your country or institution) were obtained?
    \item[] Answer: \answerNA{} % Replace by \answerYes{}, \answerNo{}, or \answerNA{}.
    \item[] Justification: The paper does not involve crowdsourcing or research with human subjects.

\end{enumerate}

\end{document}